\documentclass[journal]{IEEEtran}
\usepackage{hyperref}
\usepackage[mathscr]{eucal}
\usepackage{amsmath}
\usepackage{graphicx}
\usepackage{mathtools}
\usepackage{caption}
\usepackage{subcaption}
\usepackage{amsfonts}
\usepackage{algorithm}
\usepackage{algorithmicx} 
\usepackage{algcompatible}
\usepackage[noend]{algpseudocode}
\usepackage{algpseudocode}
\usepackage{xcolor}

\begin{document}
%
\title{Spatiotemporal tensor completion for improved urban traffic imputation}
%
%
%

\author{Ahmed~Ben Said,
        Abdelkarim~Erradi,
\thanks{A. Ben Said is with the Department of Computer Science \& Engineering, College of Engineering, Qatar University, Doha, 2713, Qatar. Email abensaid@qu.edu.qa} 
\thanks{A. Erradi is with the Department of Computer Science \& Engineering, College of Engineering, Qatar University, Doha, 2713, Qatar. Email erradi@qu.edu.qa
}}

\markboth{IEEE TRANSACTIONS ON INTELLIGENT TRANSPORTATION SYSTEMS}%
{Shell \MakeLowercase{\textit{et al.}}: Bare Demo of IEEEtran.cls for IEEE Journals}

\maketitle

\begin{abstract}
Effective   management   of   urban   traffic   is   important for any smart city initiative. Therefore, the quality of the sensory traffic data is of paramount importance.  However, like  any  sensory  data,  urban  traffic  data  are  prone  to  imperfections  leading  to  missing  measurements.  In  this  paper, we  focus  on  inter-region  traffic  data  completion.  We model the inter-region traffic as  a  spatiotemporal  tensor  that  suffers  from  missing measurements.   To   recover the missing   data,   we   propose   an enhanced  CANDECOMP/PARAFAC  (CP)  completion  approach that  considers the  urban and temporal  aspects  of  the  traffic.  To  derive the  urban  characteristics,  we  divide  the  area  of  study  into regions. Then, for each region, we compute urban feature vectors  inspired  from  biodiversity which are used  to compute the urban similarity matrix. To mine the temporal aspect, we first  conduct  an  entropy  analysis  to  determine  the  most  regular time-series.  Then,  we  conduct  a  joint  Fourier  and  correlation analysis  to  compute  its  periodicity and  construct  the  temporal  matrix.  Both urban and temporal matrices are fed into  a  modified CP-completion  objective  function.  To  solve  this  objective,  we propose   an   alternating   least   square   approach   that   operates on  the  vectorized  version  of  the  inputs.  We  conduct comprehensive comparative study with two evaluation scenarios. In  the  first  one,  we  simulate  random missing values. In the second scenario, we simulate missing values at a given area and time duration. Our results demonstrate that our approach provides effective recovering performance reaching 26\%  improvement  compared  to  state-of-art CP approaches and 35\% compared to state-of-art generative model-based approaches.


\end{abstract}


%
\IEEEpeerreviewmaketitle

\section{Introduction}
Modern smart cities are increasingly deploying Internet of Things (IoT) sensors to collect and analyze data to efficiently manage urban assets and services such as public transport, utilities, traffic monitoring and public safety.
With the widespread usage of sensors, massive urban data are continuously collected. There has been a great interest in using recent advances in data analytics to exploit these data in order to deliver better urban services and solve critical problems associated to the massive urban growth such as traffic congestion and public transport efficiency. For instance, forecasting traffic flow has been one of the most successful applications of state-of-art deep learning approaches.
The collected sensory data are inevitably prone to multiple equipment-related issues and data collection imperfections causing data loss. Such data loss can be associated to multiple causes such as GPS calibration, connectivity problem or weather conditions. Missing values have dramatic consequences as it may lead to drawing misleading conclusions and therefore wrong decisions. In terms of cost, missing values may forces the city planning authority to redo the experiment in order collect the required data, hence extra budgetary cost and time delay. At a city-wide scale, multiple sensors are deployed to continuously collect traffic data. However, it is costly and technically difficult to deploy traffic sensors across every corner of a metropolitan area, not to mention the challenge of management and maintenance. Practically, urban authority relies on few sensors deployed only on key areas. Hence, traffic data imputation is important to obtain a complete overview of the overall city traffic. Completing the missing data is critical for many tasks such as estimating  travel time and congestion-aware route planning.\newline
In this paper, we address the problem of urban traffic data completion. We propose a modified CP completion approach that takes into account the urban and time context of the traffic to drive the completion algorithm.
The paper contribution can be summarized as follows:
\begin{itemize}
    \item We model the interaction between regions in the area of study as a spatiotemporal tensor. This tensor suffers from missing data which must be recovered to get better insights about the traffic flow.
    \item The tensor captures the traffic flow and hence the interaction between regions in the time domain. 
    Our choice for tensor design considers the full traffic records. i.e. our tensor is built using all locations visited during the trip from the start to the end regions. This results in a less sparse tensor  compared to the scenario where only source and destination are used to build it as more interaction between regions are derived.
    \item We propose an urban and time aware CP completion approach. The urban characteristics of each region are taken into consideration. They include the region's richness, diversity, concentration of Points of Interest (POIs) and convenience. These characteristics are used to determine the similarity between regions which is then used to augment the CP completion with additional features.
    \item We conduct a time series analysis to determine the periodicity of the traffic pattern. This periodicity is used to construct the temporal characteristics incorporated in the CP cost function.
    \item We propose an alternating least square approach to minimize the CP cost. To address the optimization, we propose to conduct the minimization on a reshaped version of the inputs. 
    \item We provide a comprehensive comparative study with multiple completion approaches to validate the effectiveness of our proposal. Our experiments are performed on two real world datasets: T-drive taxi data \cite{t_drive1,t_drive2} from Beijing and Porto taxi \footnote{\url{www.kaggle.com/c/pkdd-15-predict-taxi-service-trajectory-i/data}}.
\end{itemize}
The rest of the paper is organized as follows: Section II discusses important related work. Section III presents some tensor calculation basics used in the proposed approach. We detail in section IV the formulation of our proposed urban-aware CP completion problem. Experiments are presented in section V. The last section concludes the paper and presents an agenda for future work.

\section{Related work}
Zhang et al. \cite{deep_st} proposed a spatiotemporal learning approach to predict the citywide crowd flows. The authors introduced inflow and outflow matrices for counting the incoming/outgoing moving objects for a given region at a given time slot. The aggregation of the two matrices are used to predict the flow at time $t_{K+1}$ given the historical traffic flow till time $t_K$. The authors proposed Deep-ST, a deep neural network architecture trained on the flow tensor. The historical flow tensors are grouped into three time horizons: recent, near and distant. A stack of Convolutional Neural Networks (CNN) is trained on each time horizon flow tensor. Deep-ST exploits additional information such as weather, and type of the day (weekend/weekday) for more accurate forecasting. Hoang et al. \cite{fccf} focused on the factors affecting crowd flows including seasonal (periodic), trend (changes in periodic patterns) and residual (instantaneous) flows. Gaussian Markov random field is used to model the seasonal and trend flows. The instantaneous flow exploits the spatiotemporal dependencies of different flows in addition to the weather data. This is achieved by applying a regression analysis where information about intra-regions and inter-regions dependence and weather condition are incorporated. Huang et al. \cite{dbn_flow} proposed a Deep Belief Network (DBN) based architecture for traffic flow prediction. DBN is effective in generating features in an unsupervised fashion. On top of the DBN, a multitask layer is incorporated for supervised traffic flow prediction. 
\newline
\indent Since most of the research works on traffic and crowd flow prediction are data-driven, it is of paramount importance to maintain clean and complete data. This research problem has been extensively studied and multiple successful approaches have been developed. For instance, Sparsity Regularized SVD (SRSVD) \cite{srsvd} approach focuses on Internet traffic matrix completion. SRSVD uses Singular Value Decomposition to find a global low rank approximation of the matrix and exploits its spatiotemporal structure by augmenting the minimization problem with two spatial and temporal matrices. The problem is then solved using Alternating Least Square (ALS) minimization. Compressive Sensing (CS) \cite{cs} is closely related to matrix completion problem. CS accurately recovers information of a sparse matrix using small subset of samples.
Roughan et al. \cite{srmf} proposed Sparsity Regularized Matrix Factorization (SRMF). This approach exploits the low rank and spatiotemporal property of the traffic matrix to estimate the missing values. SRMF seeks the global low rank approximation which is then augmented with an interpolation technique such as k-nearest neighbours to fully recover the traffic matrix. Wen et al. \cite{lmafit} addressed the computation complexity of the completion problem based on the nuclear norm which requires calculating  singular value decompositions. The authors proposed the Low-rank Matrix Fitting (LMaFit), a low complexity algorithm that is based on nonlinear successive over-relaxation approach that requires solving a linear least squares problem at each iteration. \newline 
\indent However, the aforementioned solutions for traffic matrix completion operate on the two-dimensional traffic matrices whose columns are stacked. The multi-ways nature of such matrices is unfortunately ignored. Consequently, the matrix representation is simply not enough for efficient data recovery solutions. \newline
\indent In presence of more than two dimensional data, tensor representation for data recovery has been recently investigated. Indeed, a tensor can encompass more global information compared to a matrix such as an additional third dimension representing the time. 
Long et al. \cite{review_visual} reviewed state-of-art techniques of tensor completion for visual data. The authors identified two groups of approaches based on the optimization techniques used. One sets a predefined rank and optimizes the factors of tensor decomposition while the second group minimizes the rank of the estimated tensor iteratively.
Acar et al. \cite{wopt}  proposed a CP weighted optimization algorithm (CP-WOPT). A first-order optimization is utilized to solve the weighted least squares problem. CP-WOPT has been successfully used to estimate missing data in spatiotemporal internet traffic tensor. In \cite{rals}, the authors studied the convergence of the regularized ALS for tensor decomposition. Regularization is applied to avoid overfitting. The authors proved that ALS does not always converge using the Gauss-Siedel method while the regularized ALS provides better convergence and may decrease the required number of iterations.
\newline
\indent In context of urban dynamics and mobility pattern, missing data is a common issue. Li et al \cite{review1} and Ni et al \cite{review2} surveyed state-of-art techniques for traffic data completion. In \cite{bayesian}, the authors addressed the problem of missing values in intelligent transportation system using a probabilistic framework that extends the well known bayesian approach of Salakhutdinov and Mnih \cite{salakh} to the higher order tensor. However, no urban context information is used. In \cite{wang},  the data tensor represents the interaction between regions of the area of study. The set of regions is obtained based on the traffic zones provided by the transportation authority. Each data point $r_{ijk}$ is the log transform of the number of moving objects whose start point is zone $i$ and final destination is zone $j$ departing at time $k$. The temporal dimension represents one hour-slice. For better recovery performance, the authors augmented the completion approach with urban contextual factors. These factors reflect the proportion of each type of POI at each region. Although this approach attempted to take into account the urban context, it does not consider important urban factors such as the convenience and the diversity of the region in terms of POIs. In addition, the authors studied only the mobility interaction based on the start and end regions of the urban mobility data which do not capture the instantaneous interaction between regions while  travelling from source to destination.
Tan et al. \cite{tan} proposed an algorithm that uses the multimode transport information to predict the traffic flow with a low-rank constraint. The authors also addressed the forecasting problem in the presence of missing data. However, the proposed method does not scale well with very large traffic tensor. Li et al. \cite{weakly} proposed a completion approach for tensor built using passenger flow from a metro service. The completion objective is regularized by introducing weakly dependent penalty and graph penalty and solved using Block Coordinate Descent. The main assumption is that two stations are less likely to be highly-dependent in terms of traffic flow profile. However, such assumption may not be valid for general road network traffic where the traffic flow is not restricted for just one mean of transportation.
\newline
\indent With recent advances in deep learning, data imputation has been addressed using generative models. Yoon et al. \cite{gain} proposed a Generative Adversarial Network (GAN) \cite{gan} based model in which the generator observes parts of the real data and completes the missing components. The discriminator is trained to discriminate between the observed and imputed data while being provided with a hint vector. This vector guides the discriminator to improve the quality of imputation while ensuring that the generator completes the missing information according to the data distribution. In \cite{misgan}, the authors proposed another GAN based data completion approach named MISGAN. Two discriminator-generator pairs are used, one dedicated for the mask and the other for the data. The aim is to strengthen the imputation performance by modeling the distribution of the masks responsible for missing data. In their experiments, the authors considered only the scenario of completely random missing data. Boquet et al. \cite{part-c} used Variational Autoencoder (VAE) \cite{vau} to develop an end-to-end solution for traffic forecasting which can handle data imputation. The imputation module consists of a recognition model that, once trained, can map the traffic samples to a latent space. A decoder, trained to reconstruct the traffic samples from the latent space, can then be used to generate the imputed samples. In \cite{gp-vae}, GP-VAE, a novel VAE-based technique is proposed. Gaussian process prior and Cauchy kernel are used to model the temporal dependencies of the data. Variational parameters are predicted using the inference model which takes the data with missing information. GP-VAE is validated on benchmark tasks and medical data. Mattei et al. \cite{miwae} proposed MIWAE, an Importance-Weighted Autoencoder approach dedicated for missing at random data imputation. MIWAE maximizes a lower bound of the observed log-likelihood without any additional computational overhead compared to the Importance Weighted Autoencoders (IWAE) \cite{iwae}. In \cite{not_miwae}, the authors presented not-MIWAE, the not-missing-at-random IWAE to deal with data missing not at random. A deep neural network is used to model the conditional distribution of the pattern of missing values, hence acquiring the knowledge about the type of missingness. The proposed model maximizes a lower bound of the joint likelihood and a reparameterisation trick allows deriving the stochastic gradients of the bound for the latent and data spaces. Gondara et al. \cite{mida} proposed an unsupervised approach based on overcomplete deep denoising autoencoder. At the encoder level, the number of neurons per layer increases by a factor as the model goes deeper while at the decoder layer, the number of neurons is scaled back to the original data dimensionality.
Although generative models, particularly GAN-based, have achieved state-of-art results for imputation tasks,  they are difficult to train. Indeed, they generally involve latent variables that fail to represent the data hence making the interpretation and understanding of the imputation difficult. Furthermore, due to the loss formulation, GAN models suffer from mode collapse in addition to convergence issues \cite{gan_pb}.
In this work, we address the problem of missing values in context of urban mobility data. Our approach relies on the CP completion method. More specifically, we advocate including urban and temporal information to model the spatiotemporal interaction between regions which leads to better performance in terms of traffic data recovery and imputation.

\section{Preliminaries}
We present in this section some basic preliminaries and definitions related to tensor calculation.\newline
A tensor is a multidimensional array. The order of the tensor is its number of dimensions. The zero order tensor is a scalar. A first order tensor is a vector. A second order tensor is a matrix. For more than two dimension, the general representation is the tensor. \newline
We use Euler script letter
$\mathscr{T}$  to denote a tensor of order $n \geq 3$. We use bold capital letter $(\mathbf{A,B,C})$ to denote a matrix in $\mathbb{R}^{I_1\times I_2}$ and lower case $(a,b,c)$ to denote a vector. The entry of a matrix $\mathbf{A} \in \mathbb{R}^{I_1\times I_2}$ is denoted by $a_{i_1i_2}$. The entry of a tensor $\mathscr{T} \in \mathbb{R}^{I_1\times I_2 \times ... I_n}$ is denoted by $t_{i_1i_2,...i_n}$. The nuclear norm of a tensor $\mathscr{T}$ is $||\mathscr{T}||_1 = \sum_{i_1} \sum_{i_2}... \sum_{i_n} |t_{i_1i_2...i_n}|$. The Frobenius norm of a tensor $\mathscr{T}$ is $||\mathscr{T}||_F = \Big(\sum_{i_1} \sum_{i_2}... \sum_{i_n} t^2_{i_1i_2...i_n}\Big)^{\frac{1}{2}}$.
\newline
We present in followings, some definitions related to the matrix tensor calculus.
\newline

\noindent \textbf{Definition 1.} The Hadamard product $(\mathbf{*})$ of two tensors of the same size is the element wise multiplication of its entries. Let $\mathscr{T}_1 \in \mathbb{R}^{I_1\times I_2 \times ... I_n}$ and $\mathscr{T}_2 \in \mathbb{R}^{I_1\times I_2 \times ... I_n}$ be two tensors, the Hadamard product, denoted $\mathscr{T}_1*\mathscr{T}_2$ is the tensor whose entries $(\mathscr{T}_1*\mathscr{T}_2)_{i_1i_2...i_n} = t^{(1)}_{i_1i_2...i_n}t^{(2)}_{i_1i_2...i_n}$.
\newline

\noindent \textbf{Definition 2.} The Kronecker product ($\mathbf{\otimes}$) of matrices $\mathbf{A} \in \mathbb{R}^{I_1\times I_2}$ and $\mathbf{B} \in \mathbb{R}^{I_3\times I_4}$ is a matrix $\mathbf{C} \in \mathbb{R}^{I_1I_3 \times I_2I_4}$ defined as:
\begin{equation}
\mathbf{C} = \mathbf{A} \otimes \mathbf{B}=
  \begin{bmatrix}
a_{11} \mathbf{B} & a_{12}\mathbf{B} & ...\\ 
a_{21}\mathbf{B} &a_{22}\mathbf{B}  & ... \\ 
 \vdots & \vdots  & \ddots 
\end{bmatrix}  
\end{equation}
\newline

\noindent \textbf{Definition 3.} The Khatri-Rao product $(\mathbf{\odot})$ of matrices  $\mathbf{A} \in \mathbb{R}^{I_1\times I_3}$ and $\mathbf{B} \in \mathbb{R}^{I_2\times I_3}$ is a matrix $\mathbf{C} \in \mathbb{R}^{I_1I_2\times I_3}$ defined as:
\begin{equation}
   \mathbf{C} = \mathbf{A} \odot \mathbf{B} = \Big[a_1 \otimes b_1 \; a_2 \otimes b_2\; ...\Big] 
\end{equation}
where $a_i$ and $b_j$ are the $i^{th}$ and $j^{th}$ column of $\mathbf{A}$ and $\mathbf{B}$ respectively.
\newline

\begin{figure}[h!]
  \centering
  \includegraphics[scale=0.45]{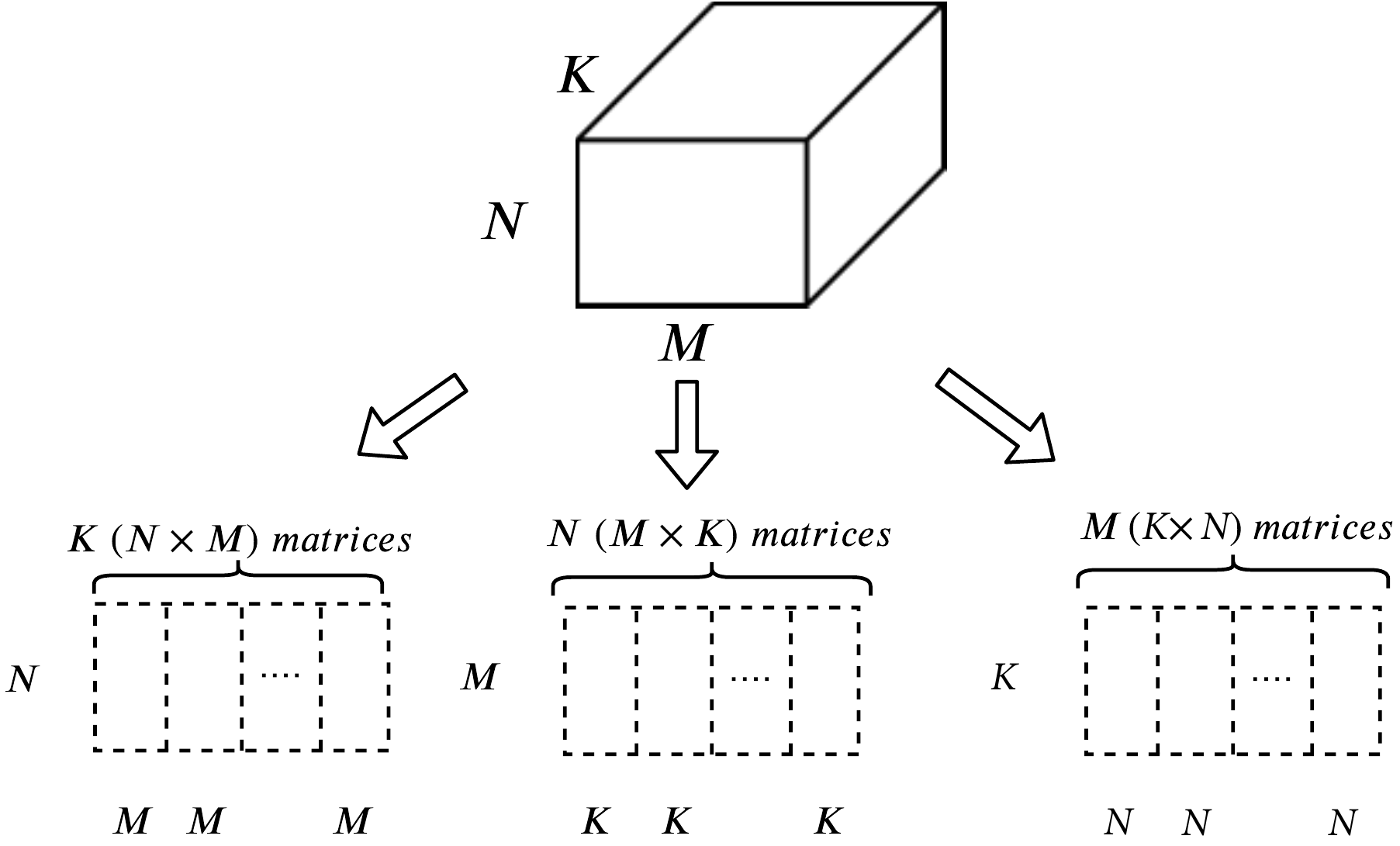}
  \caption{Matricization (Unfolding) of a 3D tensor of size ($N \times M \times K$) to three matrices of sizes ($K\times NM$), $(N\times MK)$ and ($M\times KN$).}
  \label{matricization}
\end{figure}

\noindent \textbf{Definition 4.} A mode $n$-matricization of the $N^{th}$ order tensor, known as unfolding, is the process of organizing the tensor into a matrix. We illustrate in Fig. \ref{matricization} the first, second and third order matricization of a $N\times M \times T$ tensor. The mode $n$ matricization is denoted as $T_{(n)}$.
\newline

\noindent \textbf{Definition 5.} The $n^{th}$ order tensor $\mathscr{T} \in \mathbb{R}^{I_1\times I_2 \times ... I_n}$ is rank one if it can be written as the outer products of $N$ vectors:
\begin{equation}
    \mathscr{T} = a^{(1)}\; o\; a^{(2)}\; o\; ... \;o\; a^{(N)} 
\end{equation}
$a^{(r)}$, $1\leq r \leq N$, is a vector in $\mathbb{R}^{I_r}$. $o$ is the outer product. 
\newline

\begin{figure}[h!]
  \centering
  \includegraphics[scale=0.53]{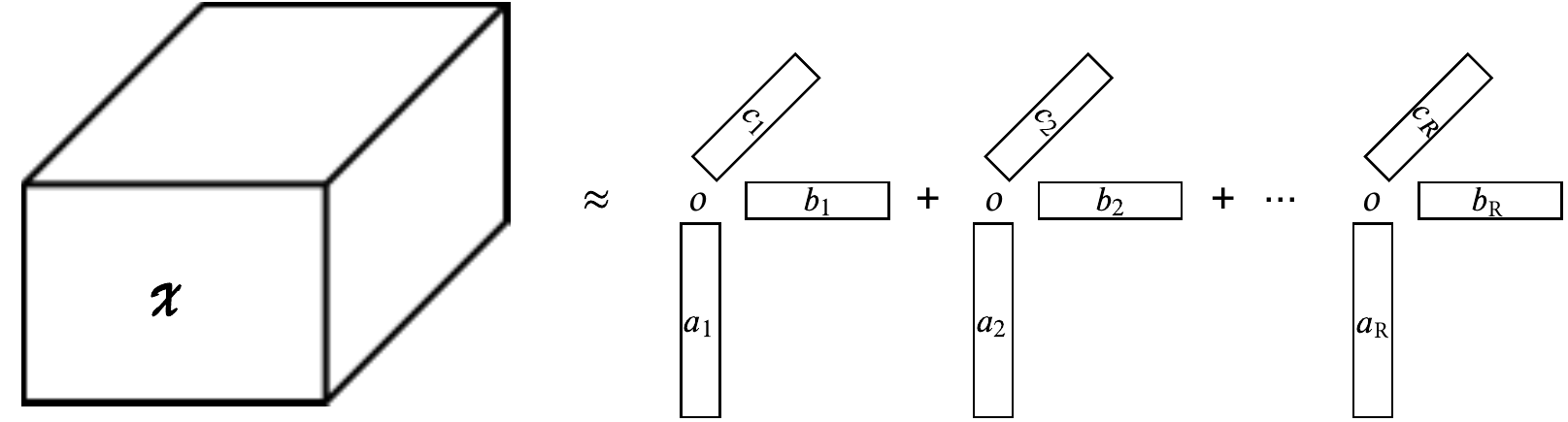}
  \caption{CP decomposition of third-order tensor.}
  \label{cp-decomposition-figure}
\end{figure}

\noindent \textbf{Definition 6.} The  CANDECOMP/PARAFAC (CP) approach decomposes the tensor as a sum of vectors from rank one components:
\begin{equation}
    \mathscr{T} = \sum\limits_{i=1}^R a^{(1)}_r\;o\;a^{(2)}_r\;o\;...\;o\;a^{(N)}_r = [\![\mathbf{A^{(1)}},...,\mathbf{A^{(N)}} ]\!]  
\end{equation}
$a^{(i)}_r$ is the the $r^{th}$ vector of matrix $A^{(i)}$ and  $[\![\;]\!]$ denotes the CP decomposition. The set of matrices $\mathbf{A^{(i)}}$ are the latent factor matrices. As an example, let $\mathscr{T} \in \mathbb{R}^{I_1\times I_2 \times I_3}$ a third-order tensor. Its CP decomposition is:
\begin{equation}
    \mathscr{T} = \sum\limits_{i=1}^R a_r\;o\;b_r\;o\;c_r
\end{equation}
The factor matrices are the vector combinations from the rank-one components: $\mathbf{A} = [a_1,a_2,...,a_R] \in \mathbb{R}^{I_1\times R} $,  $\mathbf{B} = [b_1,b_2,...,b_R] \in \mathbb{R}^{I_2\times R} $ and $\mathbf{C} = [c_1,c_2,...,c_R] \in \mathbb{R}^{I_3\times R}$.
Fig \ref{cp-decomposition-figure} illustrates the CP decomposition of a third-order tensor.

\section{Traffic flow data completion using urban and time aware CP approach}
In this section, we detail the formulation of the traffic  tensor completion using an enhanced CP approach. First, we introduce the formulation of the problem. Next, we present a summary of the overall data completion approach and we detail the enhanced CP completion for traffic flow tensor completion. Finally, we show how the spatiotemporal urban features can be integrated with CP to enhance the recovery performance of urban traffic information.

\subsection{Problem formulation}
\begin{figure}[h!]
  \centering
  \includegraphics[scale=0.35]{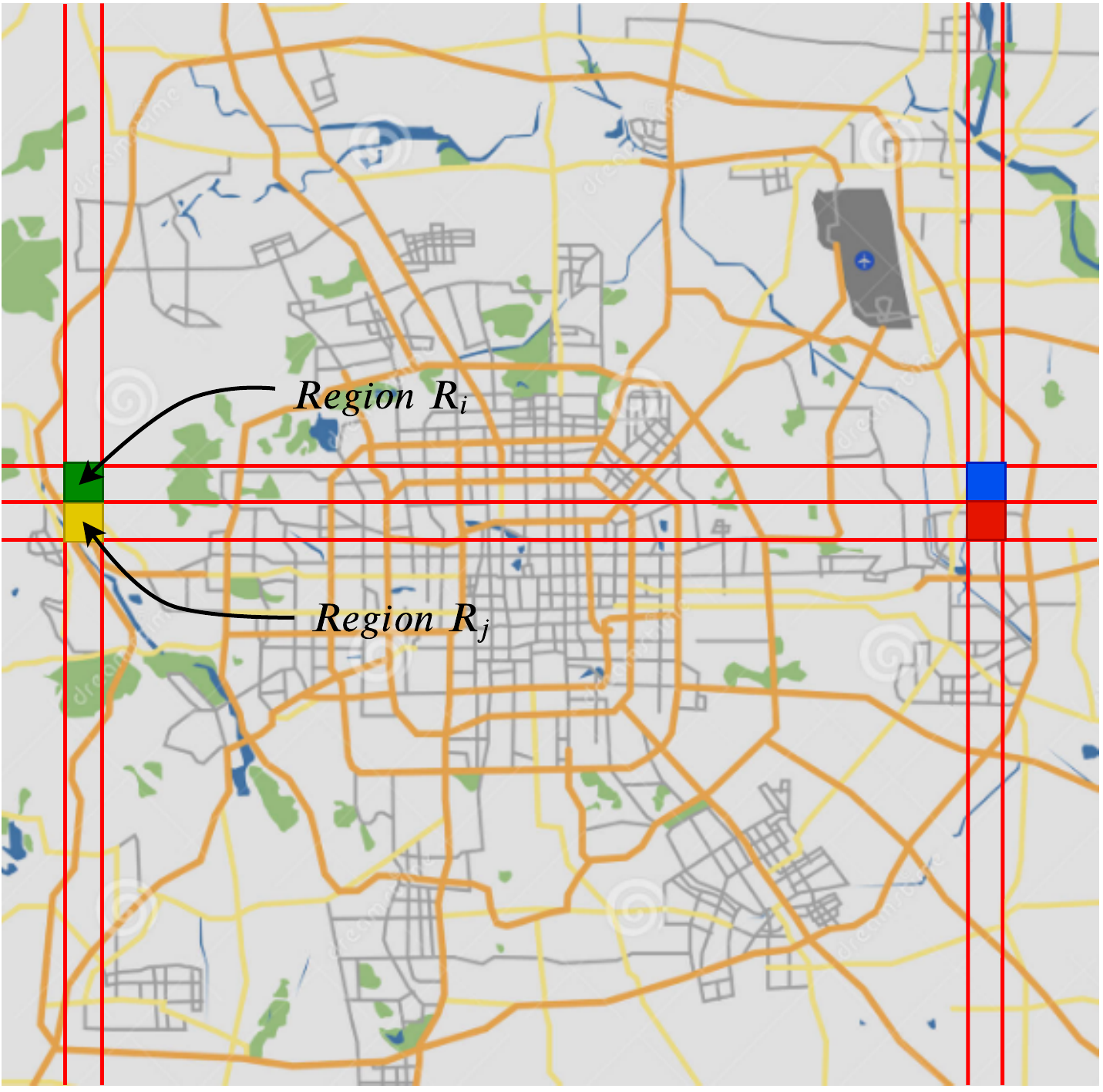}
  \caption{Grid-based segmentation of Beijing. (For clarity, we illustrate few squares)}
  \label{segmentation}
\end{figure}

Urban traffic data collected from distributed sensors are prone to multiple imperfections leading to missing measurements. To address the inter-region traffic flow data completion problem, we first segment the area of study into $M$ regions. Then we model the traffic flow from one region to another as a spatiotemporal tensor that may have missing measurements. Multiple approaches can be adopted for the region segmentation including administrative, morphology, grid and road segments based segmentation \cite{segmentation1,segmentation2}. In this paper, we simply adopt a grid based approach. Specifically, we set up a boundary box over the area of study and divide it into elementary squares as illustrated in Fig. \ref{segmentation}. The size of the elementary squares is adjustable depending on the desired granularity e.g. $1 km^2$, $2 km^2$, etc. 
Given $M$ regions and $T$ time intervals, a traffic flow tensor $\mathscr{X} \in \mathbb{R}^{M\times M \times T}$ is a third order tensor where each entry $x_{ijk}$ represents the number of objects (car, bike, pedestrians ...) located at region $R_i$ at time $k$ and relocated at region $R_j$ at time $k+1$. In previous approaches, this tensor is constructed by considering only the start and end location of the moving objects (e.g. car, bus, etc.). Thus, it does not capture the complete travel patterns of these objects and the intermediate visited segmentation squares.
Unlike these approaches, we consider every sampled location from the travel pattern  of each moving object to build the traffic flow tensor in order to provide a complete overview of the traffic during the studied time span.\newline
Let $\mathscr{W} \in \mathbb{R}^{M\times M \times T}$ be a binary tensor such that:
\[
    w_{ijk}=\left\{
                \begin{array}{ll}
                  0 \quad if\; x_{ijk}\; is\; missing\\
                  1 \quad otherwise\\
                \end{array}
              \right.
  \]
 $\mathscr{W}$ models the perturbation that leads to the missing information in the data tensor. 
The observed traffic flow tensor $\mathscr{Y}$, i.e. the tensor with missing data is the element wise product of the complete tensor $\mathscr{X}$ with the perturbation $\mathscr{W}$:
\begin{equation}
    \mathscr{Y} = \mathscr{W}*\mathscr{X}
\end{equation}
Our goal is to recover $\mathscr{X}$ given the observed tensor $\mathscr{Y}$ by seeking an approximate tensor $\hat{\mathscr{X}}$ which is as close as possible to the true and complete tensor $\mathscr{X}$. Our strategy consists of introducing a prior knowledge related to the urban and time context of the traffic flow.

\subsection{Overview of the proposed enhanced CP for traffic flow data completion}
\begin{figure*}[h!]
  \centering
  \includegraphics[scale=0.6]{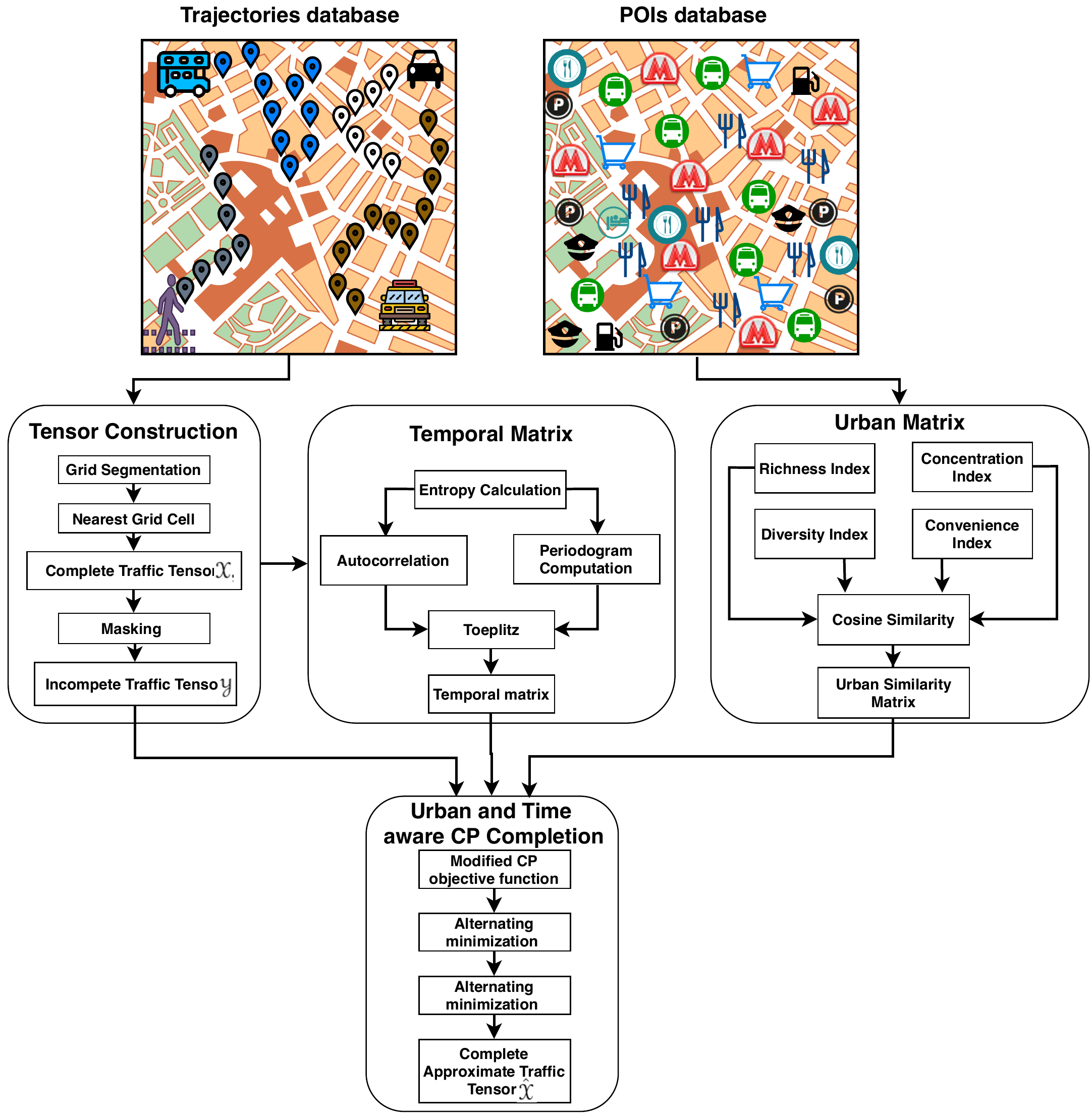}
  \caption{Traffic tensor data completion using urban and time aware CP approach.}
  \label{framework}
\end{figure*}
Fig. \ref{framework} illustrates the proposed traffic data completion approach using an enhanced CP approach  taking into account the temporal aspects of the input data and the urban characteristics of the area of study. 
Given an area of study and its associated database of moving object trajectories and POI, the CP completion approach is a four-stage process: tensor construction, temporal matrix calculation, urban matrix calculation and the urban and time aware CP completion solver. 
First, the area of study is segmented into a grid of elementary blocks. Each block is identified by its central location.
Given a set of trajectories of moving objects e.g. pedestrians, buses, taxis, each trajectory sample is associated to the nearest block with its associated timestamp to construct the traffic tensor. This tensor is corrupted resulting in missing information. Using  the POI data, we construct the urban context similarity matrix. For each block, we derive the following metrics: Richness, Diversity, Concentration and Convenience. These metrics are then used to construct the urban similarity matrix using cosine similarity. To derive the temporal matrix, we conduct an entropy analysis to determine the most regular time series in the traffic tensor. Then, we conduct a joint Fourier and correlation analysis to determine the periodicity of this particular time series. The calculated period is used to construct the temporal matrix which is a specific Toeplitz matrix. The two matrices are fed to a modified CP completion objective function which is optimized using an alternating minimization approach. The objective of the minimization is to reduce the error between an approximate traffic tensor and the true one. \newline
\indent We present, in the following, the formulation of CP completion problem and detail the calculation of the urban context and temporal matrices. Then, we present, the modified urban an temporal aware CP completion objective function and how to solve it in order to obtain the approximate complete traffic tensor.

\subsection{CP completion for traffic tensor recovery}
CP completion aims at recovering a tensor $\mathscr{X} \in \mathbb{R}^{M\times M \times T}$ given its rank $R$:
\begin{equation}
    \mathscr{X} \approx [\![\mathbf{A}, \mathbf{B},\mathbf{C} ]\!] = \sum\limits_{r=1}^R a_r\;o\;b_r\;o\;c_r
    \label{approx}
\end{equation}
The CP optimization problem can be formulated as:
\begin{equation}
    \text{minimize}\;f(A,B,C)\;=\; ||\mathscr{W}*\Big(\mathscr{X}- [\![\mathbf{A}, \mathbf{B},\mathbf{C} ]\!]\Big)||^2_F
    \label{objective}
\end{equation}
with respect to the factor matrices $\mathbf{A}$, $\mathbf{B}$ and $\mathbf{C}$. It has been shown \cite{rals} that the regularized version of the problem \ref{objective} converges faster. It is expressed as follows:
\begin{equation}
\begin{split}
    \text{minimize}\;f_{\lambda}(A,B,C)\;=\; ||\mathscr{W}*\Big(\mathscr{X}- [\![\mathbf{A}, \mathbf{B},\mathbf{C} ]\!]\Big)||^2_F \\  
     + \lambda\Big( ||\mathbf{A}||^2_F+||\textbf{B}||^2_F+||\mathbf{C}||^2_F\Big)
\end{split}
\label{reg_objective}
\end{equation}
$\lambda>0$ is a regularization parameter allowing a tradeoff between the approximation errors and the fitting error. Problem (\ref{reg_objective}) can be solved using the regularized ALS technique. Specifically, three sub-problems are derived:
\begin{equation}
    \begin{tabular}{l}
         \(\mathbf{A}^{k+1} = \underset{\check{\mathbf{A}} \in \mathbb{R}^{M \times R}}{argmin}\;\Big\|\mathbf{W_{(1)}}\Big(\mathbf{X_{(1)}}-\check{\mathbf{A}}(\mathbf{C}^k \odot \mathbf{B}^k)^T\Big)\Big\|^2_F\)\\ \(+\lambda||\check{\mathbf{A}}||^2_F\)   \\
         \(\mathbf{B}^{k+1} = \underset{\check{\mathbf{B}} \in \mathbb{R}^{M \times R}}{argmin}\;\Big\|\mathbf{W_{(2)}}\Big(\mathbf{X_{(2)}}-\check{\mathbf{B}}(\mathbf{C}^k \odot \mathbf{A}^{k+1})^T\Big)\Big\|^2_F\)\\
         \(+ \lambda||\check{\mathbf{B}}||^2_F\)\\
         \( \mathbf{C}^{k+1} = \underset{\check{\mathbf{C}} \in \mathbb{R}^{T \times R}}{argmin}\;\Big\|\mathbf{W_{(3)}}\Big(\mathbf{X_{(3)}}-\check{\mathbf{C}}(\mathbf{B}^{k+1} \odot \mathbf{A}^{k+1})^T\Big)\Big\|^2_F\)\\ 
         \(+ \lambda||\check{\mathbf{C}}||^2_F\)\\
    \end{tabular}
    \label{reg_subproblems}
\end{equation}
Where $\mathbf{W_{(i)}}$ and $\mathbf{X_{(i)}}$ are the $i^{th}$ order matricization of tensors $\mathscr{W}$ and $\mathscr{X}$ respectively, $T$ is the transpose operator and $k$ refers to the number of iterations. We can clearly notice that $\lambda||\check{\mathbf{A}}||^2_F$, $\lambda||\check{\mathbf{B}}||^2_F$ and $\lambda||\check{\mathbf{C}}||^2_F$ do not depend on $k$. It is worth noting that problem \ref{reg_objective} always converges toward a global minimum \cite{chemo}. However, the obtained optimal solution is related to the regularized problem \ref{reg_objective}, not to problem \ref{objective}. 

\subsection{Urban and time aware CP completion}\label{cp}
Moving patterns accross urban area have spatial and temporal dependencies. For example, at 5 PM, the end of working hours and in the city centers, traffic is usually slow with many pedestrians, cars, buses, etc.
In addition, traffic flow is also characterized by the so called urban context \cite{wang,urban1,urban2}, that is the characteristics of the surroundings such as presence of POIs including transportation facilities (metro, bus and subway stations ...), shopping malls, coffee shops, etc.
Wang et al. \cite{wang} attempted to incorporate urban context information in the tensor completion problem. Authors defined an urban matrix which captures the similarity between regions in term of POI categories proportion. By category, we refer to the type of POI such as shopping, transportation, restaurant, etc. In addition, temporal information is also incorporated. It reflects the intensity of moving patterns from source to destination. In \cite{srsvd,zhou}, authors modeled this information as a simple Toeplitz matrix $\mathbf{To} $ of the form:
\begin{equation}
\mathbf{To} =
  \begin{bmatrix}
1 & -1 & 0 &...\\ 
0 & 1  & -1 & \ddots \\
0 & 0 & 1 & \ddots \\
 \vdots & \ddots  & \ddots & \ddots
\end{bmatrix}  
\end{equation}
Matrix $\mathbf{To}$ is characterized  with the central diagonal of ones, and the first
upper diagonal of -1. It simply indicates that traffic flows at adjacent time slots are similar. Authors in \cite{srsvd} recommend incorporating a domain knowledge in the design of $\mathbf{To}$ such as the periodicity of the traffic data rather than assuming similarity with adjacent time slots. The authors proposed another form of temporal matrix in which the temporal similarity is offset by a period of 24h assuming diurnal patterns in the tensor.\newline
\indent We detail in the followings, our proposed urban similarity and temporal matrices which will be incorporated in the CP completion problem.
\subsubsection{Urban context similarity matrix}
In ecological and biogeographical studies, statistical measures have been established to characterize the diversity of an area in terms of species. This allows to obtain a quantitative estimate of the biological diversity. Inspired by the same concept, an area is characterized by its POIs diversity. For this, we use the study in \cite{poi}. Specifically, it considers Hill numbers as a measurement of POIs diversity. They are multifaceted measurements of order $q$ and expressed as:
\begin{equation}
    \prescript{q}{}{D} = \Big(\sum\limits_{i=1}^s p_i^q\Big)^{\frac{1}{1-q}}
    \label{hill}
\end{equation}
where $p_i$ is the $i^{th}$ POI category proportion, $s$ is the number of POIs categories and $q$ is the order. From Hill numbers, we define the following measures for each region $R_i$:
\newline

\noindent \textbf{Definition 7.} For $q =0$, we define the Richness index $Rch$:
\begin{equation}
    Rch = \prescript{0}{}{D} = \Big(\sum\limits_{i=1}^s p_i^0\Big)^{1} = s
\end{equation}
In other words, the richness index $Rch$ is the number of POIs categories at region $R_i$. Therefore, the presence of higher number of POIs categories indicates a richer region. We note that $Rch$ does not depend on the number of POIs of each category.
\newline

\noindent \textbf{Definition 8.} For $q=1$, Eq. \ref{hill} is not defined. However, its limit when $q \rightarrow 1$ is the exponential of the Shannon index. We define the Shannon diversity index $Sh$ as: 
\begin{equation}
Sh = \prescript{1}{}{D} = \underset{q \rightarrow 1}{lim} \prescript{q}{}{D} = exp\Big(-\sum\limits_{i=1}^s p_i\; log\;(p_i)\Big) 
\end{equation}
$Sh$ expresses the amount of randomness in the POIs categories and the number of POIs. Lower entropy values indicates greater randomness and vice versa.
\newline

\noindent \textbf{Definition 9.}
For $q=2$, we define the region concentration $Ctr$:
\begin{equation}
    Ctr = \prescript{2}{}{D} = 1/\Big(\sum\limits_{i=1}^s p_i^2\Big)
\end{equation}
It is the inverse of the Simpson index which reflects the probability that two sampled elements randomly drawn from large community would belong to the same categories. 
\newline

\noindent \textbf{Definition 10.} For a region $R_i$, we define its traffic convenience $Co$, that is the proportion of POIs associated to transportation category such as public transport stations, parking lots, etc.
\newline

\noindent \textbf{Definition 11.} For each region $R_i$, we define its urban characteristic vector $v$:
\begin{equation}
    v = [Rch,Sh,Ctr,Co]
\end{equation}
The urban context similarity matrix $\mathbf{U}$ is a matrix whose element $u_{ij}$ reflets the urban similarity between region $R_i$ and $R_j$ that is:
\begin{equation}
    u_{ij} = \frac{v_{R_i}\cdot v_{R_j}}{||v_{R_i}||\;||v_{R_j}||}
\end{equation}
Each element is the cosine similarity between each pair of regions. It ranges between -1 and 1. A value closer to 1 indicates  high similarity. The urban similarity matrix is driven by the richness, convenience, concentration and diversity indexes. When planning a trip from source to destination, one usually avoids crowded areas with high congestion and concentration of POIs. Such information is  captured by the proposed design of the urban similarity matrix.
\subsubsection{Temporal similarity matrix}
As  highlighted in section \ref{cp}, temporal dependency information is usually manifested across adjacent timestamps or offset by the period of the traffic data. However, simply assuming a 24h period of traffic is inconsistent particularly with presence of missing data as it is problematic to detect the periodicity. Furthermore, traffic patterns are not consistently regular every 24h as  the traffic significantly changes from weekdays to weekend and can easily be disrupted by any disturbance on the road network.\newline
To overcome this issue, we adopt a time series analysis strategy to construct the temporal similarity matrix. More specifically, we consider the traffic tensor from time series perspective, that is the data across the third dimension of the tensor $\mathscr{Y}$. Then, we determine the most regular time series as it will provide the most accurate periodicity estimate. A joint robust Fourier and autocorrelation is conducted to determine the period. This period is then used to construct the temporal matrix using the Toeplitz form.
\newline

\noindent \textbf{Definition 12.} A traffic time series $ts_{ij:}$ represents the traffic information of pair of regions $R_i$ and $R_j$ across the full time horizon $T$. In the remainder, we refer to $ts_{ij:}$ as simply $ts$.
\newline 

\noindent \textbf{Definition 13.} The most regular time series $ts$ is defined as the one with the least Sample Entropy (SamEn) \cite{sampEn} value. Sample Entropy has been widely used for time series analysis. It calculates the irregularity and reflects the randomness and complexity of the time series. The lower the value of SampEn, the more regular the time series is. Given a time series $ts = {ts_1,ts_2,..., t_L}$ and a template vector $ts^m$ of length $m$ from $ts$ where $ts^m_i=\{ts_i,ts_{i+1},...,ts_{i+m-1}\}$, the distance function between two template vectors is:
\begin{equation}
    d(m,i,j) = d[ts^m_i,ts^m_j] = \underset{k = 1...m}{max}\Big\{ \big| ts_{i+k-1}-ts_{j+k-1} \big| \Big\} 
\end{equation}
Let $\Theta^m_i(r)$ be the number of template vectors within distance less or equal a threshold $th$ from $ts^m_i$ is:
\begin{equation}
    \Theta^m_i(r) = \sum\limits_{j=1,j \neq i}^{N-m}\Omega(m,i,j,th)
\end{equation}
where:
\[
    \Omega(m,i,j,th)=\left\{
                \begin{array}{ll}
                  1 \quad if\; d(m,i,j) \leq th\\
                  0 \quad otherwise\\
                \end{array}
              \right.
  \]
The probability that two template vectors of length $m$ will match is defined as:
\begin{equation}
    \Delta^m_{th} = \frac{1}{N-m}\sum\limits_{i=1}^{N-m}  \Theta^m_i(r)  
\end{equation}
Finally, SamEn is expressed as:
\begin{equation}
    SampEn(m,th,N) = ln\Big(\frac{\Delta^m_{th}}{\Delta^{m+1}_{th}}\Big)
\end{equation}
SampEn cannot be directly applied as the time series contains missing values. To solve this problem, we adopt the strategy proposed in \cite{KeepSampEn} named KeepSampEn, that is the template vector $ts^m$ must not contain any missing values. Such straightforward approach has shown great stability performance and robustness against missing values.
\newline
\indent After identifying the most regular time series, we apply a joint Fourier and autocorrelation analysis to determine its periodicity. The approach consists of transforming the time series into frequency domain, determining the most dominant frequency and then mapping back to the time domain to calculate the period. First, in order to determine the most dominant frequency, we calculate the periodogram of the time series. It is the square of each coefficient of the Fourier Transform of $ts$. To mitigate the effect of missing data, we use the Lomb-Scargle periodogram \cite{lomb}. This periodogram is widely used in astronomy where missing data is a common issue. At a frequency $f_k$, it is defined as:
\begin{equation}
\begin{split}
        P(f_k) = \frac{1}{2}\Bigg(\frac{\sum_{i=1}^L \Big(ts[i]\:cos(2\pi f_k(ts[i]-\tau))\Big)^2}{\sum_{i=1}^L cos^2(2 \pi f_k(ts[i]-\tau)) }+ \\
        \frac{\sum_{i=1}^L \Big(ts[i]\:sin(2\pi f_k(ts[i]-\tau))\Big)^2}{\sum_{i=1}^L sin^2(2 \pi f_k(ts[i]-\tau)) } \Bigg)
\end{split}
\end{equation}
Where:
\begin{equation}
    \tau = \frac{1}{4\pi f_k}tan^{-1}\Bigg( \frac{\sum_{i=1}^L sin(4\pi f_k)}{\sum_{i=1}^L cos(4\pi f_k)}\Bigg)
\end{equation}
In this particular periodogram, the sine and cosine coefficients are separately normalized by a time constant which depends on the frequency $f_k$ in order to make the transform insensitive to time shift. To identify the most dominant frequency, we use a thresholding approach. Each coefficient is however mapped in the time domain to a period range $[\frac{N}{k} \frac{N}{k-1})$. To accurately determine the period, we use the circular autocorrelation.
Given a sequence $ts$, its circular autocorrelation is expressed as:
\begin{equation}
    Corr(\theta) = \frac{1}{N}\sum\limits_{i=1}^N ts[i]ts[i+\theta]
\end{equation}
Therefore, given a time range $[t_1,t_2)$ obtained using Lomb-Scargle periodogram, we look for the presence of peak in $\{Corr(t_1), Corr(t_1+1),...,Corr(t_2-1)\}$ using quadratic fitting. If the obtained fitting is concave, it indicates the presence of a period $t^* = \underset{t_1\leq t<t_2}{argmax}\;Corr(t)$.
Once $t^*$ is determined, we define the temporal similarity matrix as a Toeplitz matrix in which the difference is offset with $t^*$:
\begin{equation}
\mathbf{T_o} =
  \begin{bmatrix}
1 & 0 & \ldots  & -1 & 0 &\ldots\\ 
0 & 1  & \ddots & \ddots & \ddots & \ddots \\
0 & 0 & 1 & \ddots & \ddots & \ddots \\
 \vdots & \ddots  & \ddots & \ddots & \ddots & \ddots
\end{bmatrix}  
\end{equation}
\subsubsection{The optimization algorithm}
By introducing the urban and temporal contexts into the CP completion, the modified objective function is expressed as follows:
\begin{equation}
\begin{split}
    \text{minimize}\;f^u_{\lambda}(\mathbf{A},\mathbf{B},\mathbf{C})\;=\; ||\mathscr{W}*\Big(\mathscr{X}- [\![\mathbf{A}, \mathbf{B},\mathbf{C} ]\!]\Big)||^2_F \\  
     + \lambda\Big( ||\mathbf{A}||^2_F+||\mathbf{B}||^2_F+||\mathbf{C}||^2_F\Big) + \beta \Big( \left\|[\![\mathbf{UA}, \mathbf{B},\mathbf{C} ]\!]\right\|^2 \\ 
     + \left\|[\![\mathbf{A}, \mathbf{UB},\mathbf{C} ]\!]\right\|^2 + \left\|[\![\mathbf{A}, \mathbf{B},\mathbf{T_oC} ]\!]\right\|^2 \Big) 
\end{split}
\label{final_objective}
\end{equation}
Where $\beta$ is a positive regularization parameter.
In the above design of $f^u_{\lambda}$, two insights are exploited. First, the traffic data with its periodicity and temporal stability are included. Second, the urban similarity matrix reflects how one region is similar to another one in terms of POIs. At a given time, regions with high similarity are highly likely to exhibit the same traffic pattern. This knowledge is incorporated in the modified traffic tensor completion problem. Such design has been used for internet traffic data completion \cite{zhou} with different space and time context, and shown effective recovery performance.
To solve the objective \ref{final_objective}, we adopt an alternating least square procedure. First, we fix $\mathbf{B}$ and $\mathbf{C}$ and we solve for $\mathbf{A}$. Next, we fix $\mathbf{A}$ and $\mathbf{C}$ and solve for $\mathbf{B}$. Finally, we fix $\mathbf{A}$ and $\mathbf{B}$ and solve for $\mathbf{C}$. When fixing two parameters and solving for the third, the problem becomes a simple linear least squares. For example, assuming $\mathbf{B}$ and $\mathbf{C}$ are fixed, the obtained least squares problem can be expressed as follows \cite{zhou}:
\begin{flalign}
    \begin{split}
        ||\mathbf{W}_{(1)}*\Big(\mathbf{X}_{(1)}-\mathbf{A}(\mathbf{C}\odot \mathbf{B})^T\Big)||^2_F + \lambda ||\mathbf{A}||^2_F \\
        + \beta \Big( ||(\mathbf{U}\mathbf{A} (\mathbf{C}\odot \mathbf{B})^T)  ||^2_F +   ||\mathbf{A}(\mathbf{C} \odot (\mathbf{U}\mathbf{B}))^T||^2_F \\
        + ||\mathbf{A}((\mathbf{T_o}\mathbf{C})\odot \mathbf{B})^T||^2_F\Big)
    \end{split}
    \label{subproblem_a}
\end{flalign}
By writing:
\begin{equation}
    \Psi_1 = \mathbf{C} \odot \mathbf{B} \quad \Phi_1=\mathbf{C}\odot(\mathbf{U}\mathbf{B}) \quad \Gamma_1= (\mathbf{T_o}\mathbf{C}) \odot \mathbf{B}
\end{equation}
and taking the derivative of Eq. \ref{subproblem_a} with respect to $\mathbf{A}$ and setting it equal to zero, we have:
\begin{equation}
    \begin{split}
        &(\mathbf{W}_{(1)}*\mathbf{W}_{(1)}*(\mathbf{A}\Psi_1^T))\Psi_1 + \lambda \mathbf{A}\Big(\mathbf{I}_{[A]} + \Phi_1^T\Phi_1+ \Gamma_1^T\Gamma_1 \Big) + \\
        &\beta \mathbf{U}^T\mathbf{U}\mathbf{A}\Psi_1^T\Psi_1=\mathbf{W}_{(1)}*\mathbf{X}_{(1)}
    \end{split}
\end{equation}
where $I_{[A]}$ is the identity matrix whose size is the number of rows of $A$.
Let $vec(\:)$ be the operator which creates a column vector from a matrix by stacking its columns one below another:
\begin{align}
    vec(X) &= \begin{bmatrix}
           x_{1} \\
           x_{2} \\
           \vdots \\
           x_{m}
         \end{bmatrix}
\end{align}
where $x_i$ is the i$^{th}$ column of matrix $X$.
We use the following formulas:
\begin{equation}
    \begin{split}    
    &vec(\mathbf{A}\mathbf{X}\mathbf{B}) = (\mathbf{B}^T \otimes \mathbf{A})\;vec(\mathbf{X})\\
    &vec(\mathbf{A}\mathbf{B})= (\mathbf{B}^T \otimes \mathbf{I}_{[A]})\;vec(\mathbf{A}) = (\mathbf{I}_{[B^T]} \otimes \mathbf{A})\;vec(\mathbf{B})\\
    & vec(\mathbf{A})*vec(\mathbf{B}) = diag\Big((vec(\mathbf{A})\Big)\;vec(\mathbf{B})\\
    & vec(\mathbf{A}*\mathbf{B}) = vec(\mathbf{A})*vec(\mathbf{B}) = vec(\mathbf{B})*vec(\mathbf{A})
    \end{split}
\end{equation}
where $diag(x)$ is a diagonal matrix with the elements of the vector $x$ are in its diagonal.\newline
By applying the vec operator, we have:
\begin{equation}
    \begin{tabular}{l}
         \((\Psi_1^T \otimes \mathbf{I}_{[\mathbf{W}_{(1)}]})\;diag\Big(vec(\mathbf{W}_{(1)})*vec(\mathbf{W}_{(1)})\Big) \cdot\) \\
         \((\Psi_1 \otimes \mathbf{I}_{[A]})\;vec(A) + \beta \Big((\Psi_1^T\Psi_1)\otimes (\mathbf{U}^T\mathbf{U})\Big)\;vec(\mathbf{A})\)  \\ 
         \(\lambda\Big(\big(\mathbf{I}_{[A]} + \Phi_1^T\Phi_1+ \Gamma_1^T\Gamma_1 \big)^T \otimes \mathbf{I}_{[A]}\Big)\;vec(A)\)\\
         \(=vec(\mathbf{W}_{(1)})*vec(\mathbf{X}_{(1)})\)\\
         \(\Bigg((\Psi_1^T \otimes \mathbf{I}_{[\mathbf{W}_{(1)}]})\;diag\Big(vec(\mathbf{W}_{(1)})*vec(\mathbf{W}_{(1)})\Big)\)\\
         \((\Psi_1 \otimes \mathbf{I}_{[A]})+ \lambda\Big(\mathbf{I}_{[A]} + \Phi_1^T\Phi_1+ \Gamma_1^T\Gamma_1 \Big)^T \otimes \mathbf{I}_{[A]} +\) \\
         \(\beta (\Psi_1^T\Psi_1)\otimes (\mathbf{U}^T\mathbf{U})  \Bigg)\;vec(\mathbf{A})=\Delta\;vec(\mathbf{A})\)\\
         \(=vec(\mathbf{W}_{(1)})*vec(\mathbf{X}_{(1)})\)
    \end{tabular}
\end{equation}
Finally, we have:
\begin{equation}
    \begin{split}
    &\Delta\;vec(\mathbf{A}) =vec(\mathbf{W}_{(1)})*vec(\mathbf{X}_{(1)}) \\  
    & vec(\mathbf{A}) = \Big(\Delta\Big)^+ vec(\mathbf{W}_{(1)})*vec(\mathbf{X}_{(1)})
    \end{split}
    \label{solve_A}
\end{equation}
Where $\Big(\cdot \Big)^+$ is the Moore-Penrose inverse. \newline
Similarly, to solve for $\mathbf{B}$, we  fix $\mathbf{A}$ and $\mathbf{C}$, the obtained least square problems is:
\begin{flalign}
    \begin{split}
        ||\mathbf{W}_{(2)}*\Big(\mathbf{X}_{(2)}-\mathbf{B}(\mathbf{C}\odot \mathbf{A})^T\Big)||^2_F + \lambda ||\mathbf{B}||^2_F \\
        + \beta \Big( ||(\mathbf{B} (\mathbf{C}\odot (\mathbf{U}\mathbf{A}))^T)  ||^2_F +   ||\mathbf{U}\mathbf{B}(\mathbf{C} \odot \mathbf{A})^T||^2_F \\
        + ||\mathbf{B}((\mathbf{T_o}\mathbf{C})\odot \mathbf{A})^T||^2_F\Big)
    \end{split}
    \label{subproblem_b}
\end{flalign}
Let:
\begin{equation}
    \Psi_2 = \mathbf{C}\odot \mathbf{A} \quad \Phi_2=\mathbf{C}\odot (\mathbf{UA})\quad \Gamma_2 = (\mathbf{T_oC})\odot \mathbf{A} 
\end{equation}
We have:
\begin{equation}
    \begin{tabular}{c}
        \(\Bigg((\Psi_2^T \otimes \mathbf{I}_{[\mathbf{W}_{(2)}]})\;diag\Big(vec(\mathbf{W}_{(2)})*vec(\mathbf{W}_{(2)})\Big)\)   \\
        \((\Psi_2 \otimes \mathbf{I}_{[B]}) +\lambda \Big(\mathbf{I}_{[B]} + \Phi_2^T\Phi_2+ \Gamma_2^T\Gamma_2 \Big)^T \otimes \mathbf{I}_{[B]}+\) \\
         \(\beta (\Psi_2^T\Psi_2)\otimes (\mathbf{U}^T\mathbf{U}) \Bigg)\;vec(\mathbf{B})=\Delta_2\;vec(\mathbf{B})\)\\
         \(=vec(\mathbf{W}_{(2)})*vec(\mathbf{X}_{(2)})\)
    \end{tabular}
    \label{solve_B}
\end{equation}
Therefore:
\begin{equation}
    vec(\mathbf{B})= \Big(\Delta_2\Big)^+ vec(\mathbf{W}_{(2)})*vec(\mathbf{X}_{(2)})
    \label{solve_B}
\end{equation}
Finally, to solve for $\mathbf{C}$, we  fix $\mathbf{A}$ and $\mathbf{B}$. The obtained least square problems is:
\begin{flalign}
    \begin{split}
        ||\mathbf{W}_{(3)}*\Big(\mathbf{X}_{(3)}-\mathbf{C}(\mathbf{B}\odot \mathbf{A})^T\Big)||^2_F + \lambda ||\mathbf{C}||^2_F \\
        + \beta \Big( ||(\mathbf{C} (\mathbf{B}\odot (\mathbf{U}\mathbf{A}))^T)  ||^2_F +   ||\mathbf{C}(\textbf{U}\mathbf{B} \odot \mathbf{A})^T||^2_F \\
        + ||\mathbf{T_o}\mathbf{C}(\mathbf{B}\odot \mathbf{A})^T||^2_F\Big)
    \end{split}
    \label{subproblem_c}
\end{flalign}
\begin{equation}
    \Psi_3 = \mathbf{B}\odot \mathbf{A} \quad \Phi_3=\mathbf{B}\odot (\mathbf{UA}) \quad \Gamma_3 = (\mathbf{UB})\odot \mathbf{A} 
\end{equation}
We have:
\begin{equation}
    \begin{tabular}{c}
    \(\Bigg((\Psi_3^T \otimes \mathbf{I}_{[\mathbf{W}_{(3)}]})\;diag\Big(vec(\mathbf{W}_{(3)})*vec(\mathbf{W}_{(3)})\Big) \)\\
    \((\Psi_3 \otimes \mathbf{I}_{[C]}) + \lambda \Big(\mathbf{I}_{[C]} + \Phi_3^T\Phi_3+ \Gamma_3^T\Gamma_3 \Big)^T \otimes \mathbf{I}_{[C]} + \)\\
    \(\beta (\Psi_3^T\Psi_3)\otimes (\mathbf{U}^T\mathbf{U}) \Bigg)\;vec(\mathbf{C})=\Delta_3\;vec(\mathbf{C})\)\\
    \(=vec(\mathbf{W}_{(3)})*vec(\mathbf{X}_{(3)})\)
    \end{tabular}
\end{equation}
Therefore:
\begin{equation}
    vec(\mathbf{C})= \Big(\Delta_3\Big)^+ vec(\mathbf{W}_{(3)})*vec(\mathbf{X}_{(3)})
    \label{solve_C}
\end{equation}
By applying \textbf{unvec()}, the inverse vec operator, we obtain the solution to the subproblems \ref{subproblem_a}, \ref{subproblem_b} and \ref{subproblem_c}. We present in Algorithm \ref{algo}, the pseudocode of the proposed urban and time aware CP tensor completion.
\begin{algorithm}[!h]
	\caption{Urban and time aware tensor completion}
	\label{algo}
	\begin{algorithmic}[1]
		\State \textbf{Input: $\mathscr{Y}$, $\mathscr{W}$, $\mathbf{F}$, $\mathbf{T_o}$, $R$, $\beta$, $\lambda$, $tol$} 
		\State \textbf{Output:} $\mathscr{\hat{X}}$,
		\State \textbf{Initialize:}$\mathbf{A} \in \mathbb{R}^{M\times R}$, $\mathbf{B} \in \mathbb{R}^{M\times R}$ and $\mathbf{C} \in \mathbb{R}^{M\times R}$
		\State $Eval_0 = f^u_{\lambda}(\mathbf{A},\mathbf{B},\mathbf{C})$ (Eq. \ref{final_objective})
		\State \textbf{Repeat:}
		\State \quad Solve for $\mathbf{A}$ using Eq. \ref{solve_A}
		\State \quad Solve for $\mathbf{B}$ using Eq. \ref{solve_B}
		\State \quad Solve for $\mathbf{C}$ using Eq. \ref{solve_C}
		\State  \quad $Eval = f^u_{\lambda}(\mathbf{A},\mathbf{B},\mathbf{C})$  (Eq. \ref{final_objective})
		\State \quad $\epsilon = Eval_0 - Eval$
		\State \quad $Eval_0 = Eval$
		\State \textbf{Until} $\epsilon<tol$
		\State \textbf{Output}: $\mathscr{\hat{X}}$ (Eq. \ref{approx})
	\end{algorithmic}
\end{algorithm}

\section{Experimental results}
In this section, we validate the effectiveness of our completion approach. We conduct a set of experiments on two traffic datasets and compare the recovery performance with multiple state-of-art approaches.
\subsection{Data}
The data we use in our experiments are road traffic records of taxi from two cities: Porto, Portugal and Beijing, China. 
\begin{itemize}
    \item Porto Taxi: the data contain 442 trajectories of taxi cabs in Porto, Portugal. For each taxi, time stamped geolocations along with metadata are provided. After segmenting the area of study into 1$km^2$ cells and aggregating the traffic in each grid cell, we obtain a $(91\times 91\times 2880)$ traffic tensor.
    \item T-drive: The data contains 15 million time stamped GPS records of 10357 taxis from February 2$^{nd}$ to February 8$^{th}$ 2008 from Beijing, China. The average sampling rate is about 177 seconds.  The data are proprocessed to eliminate noisy records. By applying grid segmentation using 2 $km^2$ and traffic record aggregation, we obtain a tensor of size $(1516\times 1516\times 352)$.
\end{itemize}
\subsection{Methodology}
Wet set up two evaluation protocols. In the first one, given the traffic tensor, we  drop measurement at random. This is achieved by randomly generating the binary mask $\mathscr{W}$ and multiply it by the traffic tensor $\mathscr{X}$ to create the observed data $\mathscr{Y}$. However, in a realistic case, missing data are the results of a failure usually related to sensor or transmission equipment dysfunction and for a some duration. We also simulate this structured missing values scenario by imputing measurements at random cells for a time duration. For each scenario, we vary the rank parameter $R$ and the missing value rate then report the Relative Error (RE):
\begin{equation}
    RE = \frac{||\mathscr{X}-\hat{\mathscr{X}}||^2}{||\mathscr{X}||^2}
\end{equation}
Where $\mathscr{X}$ and $\hat{\mathscr{X}}$ are the true and recovered traffic tensors. We evaluate the proposed completion approach against: CP\_ALS \cite{als}, CP\_ARLS \cite{rals}, CP\_OPT \cite{opt}, CP\_WOPT \cite{wopt}, CP\_APR \cite{apr} with two configurations: row subproblems by projected quasi-Newton CP\_APR\_PQNR and row subproblems by projected damped Hessian CP\_APR\_PDNR and GCP\_OPT \cite{gcp}. We set $m=3$ and $th=0.3$ for the Sample Entropy. 
\newline 
We also evaluate the proposed approach against GAN-based approaches: GAIN \cite{gain}, MIDA \cite{mida} (for both random and structured missing values), MIWAE (for random missing values) \cite{miwae} and not-MIWAE \cite{not_miwae} (for structured missing values).
For each approach, we report the best obtained result after multiple runs.
\subsection{Recovering traffic tensor with random missing values}
\begin{figure*}[h!]
     \centering
     \begin{subfigure}[b]{.3\textwidth}
         \centering
         \includegraphics[scale=0.32]{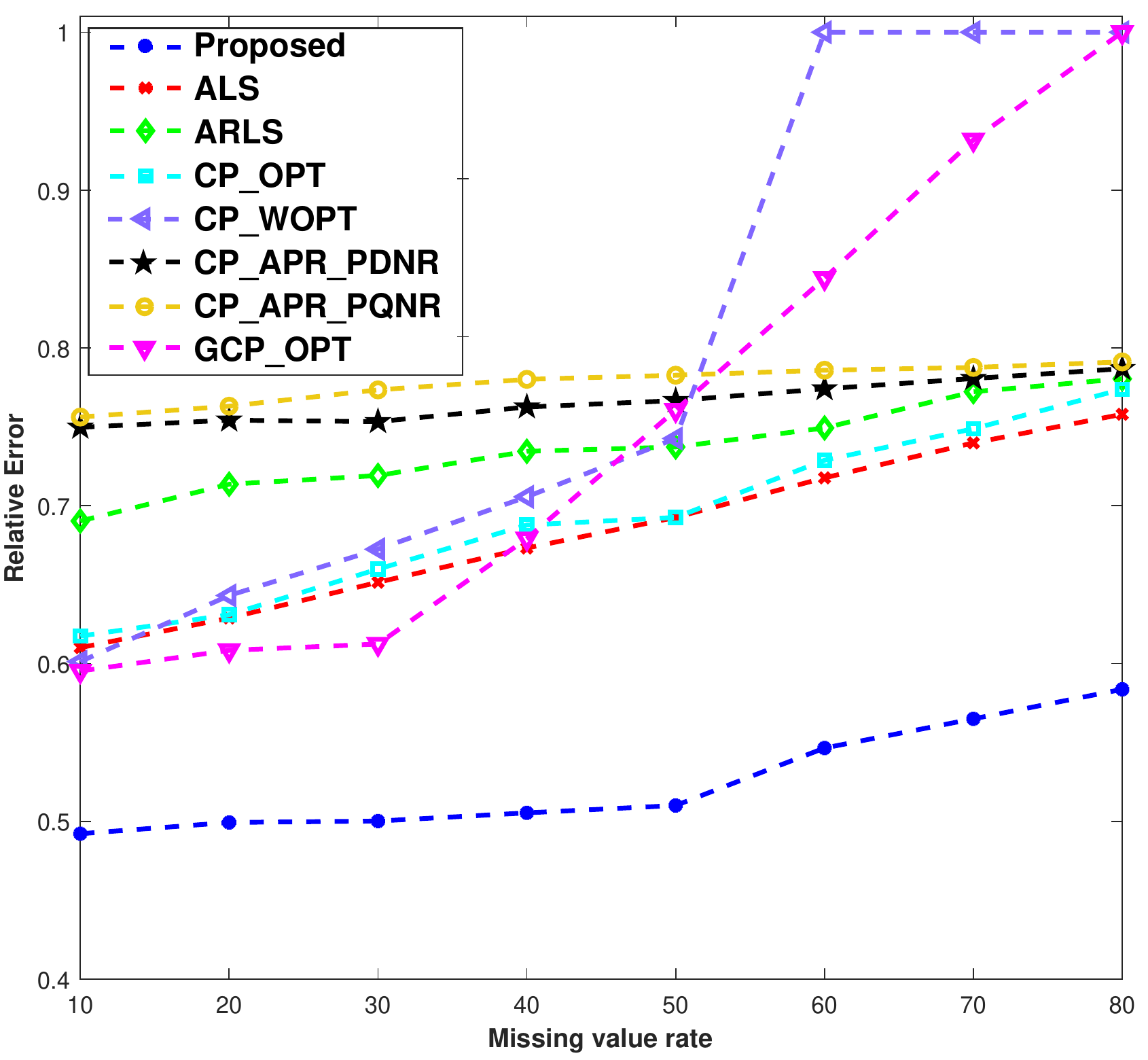}
         \caption{R=3}
         \label{porto_random_r_3}
     \end{subfigure}
     \hfill
     \begin{subfigure}[b]{0.33\textwidth}
         \centering
         \includegraphics[scale=0.32]{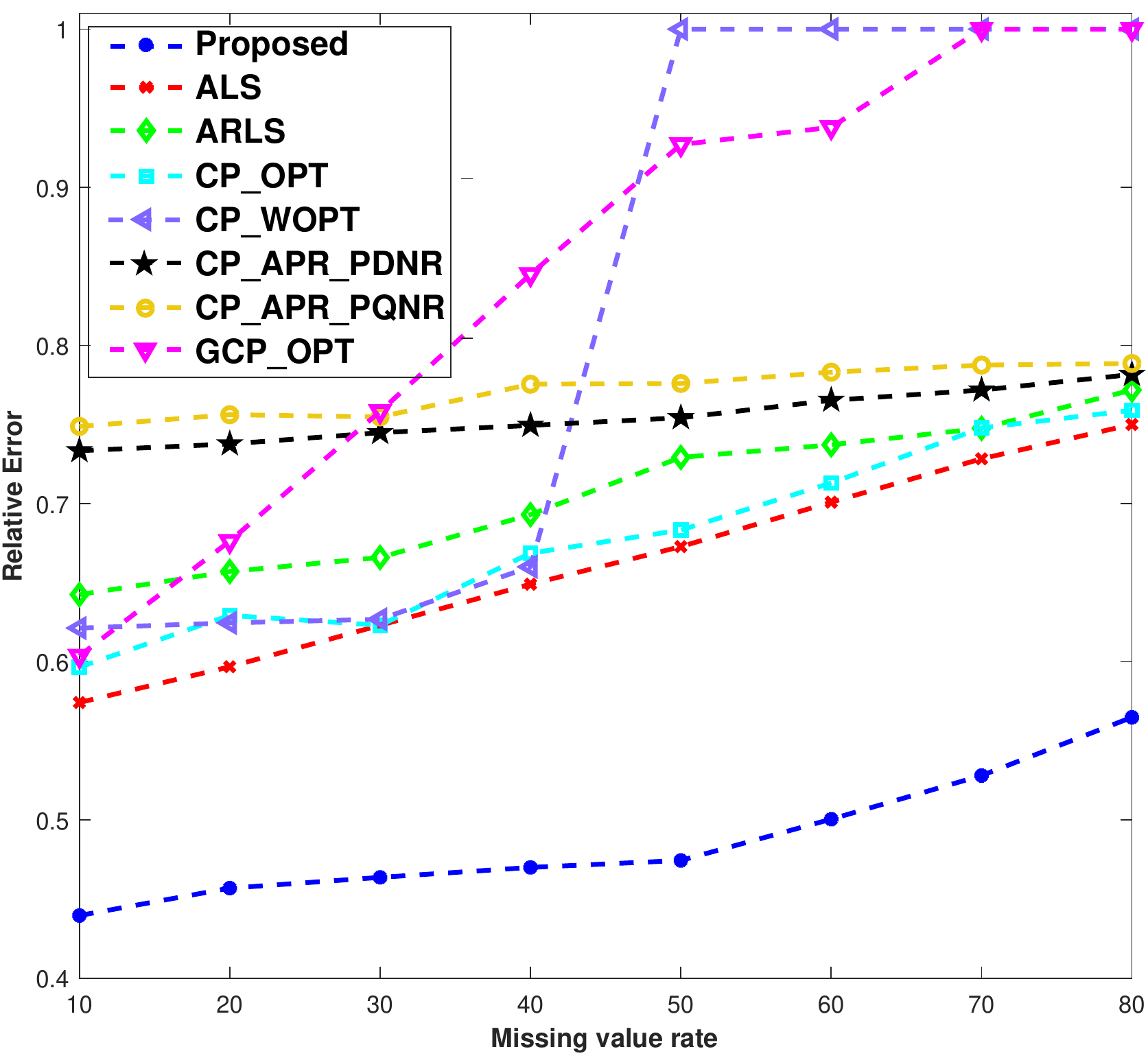}
         \caption{R=4}
         \label{porto_random_r_4}
     \end{subfigure}
     \hfill
     \begin{subfigure}[b]{0.33\textwidth}
         \centering
         \includegraphics[scale=0.32]{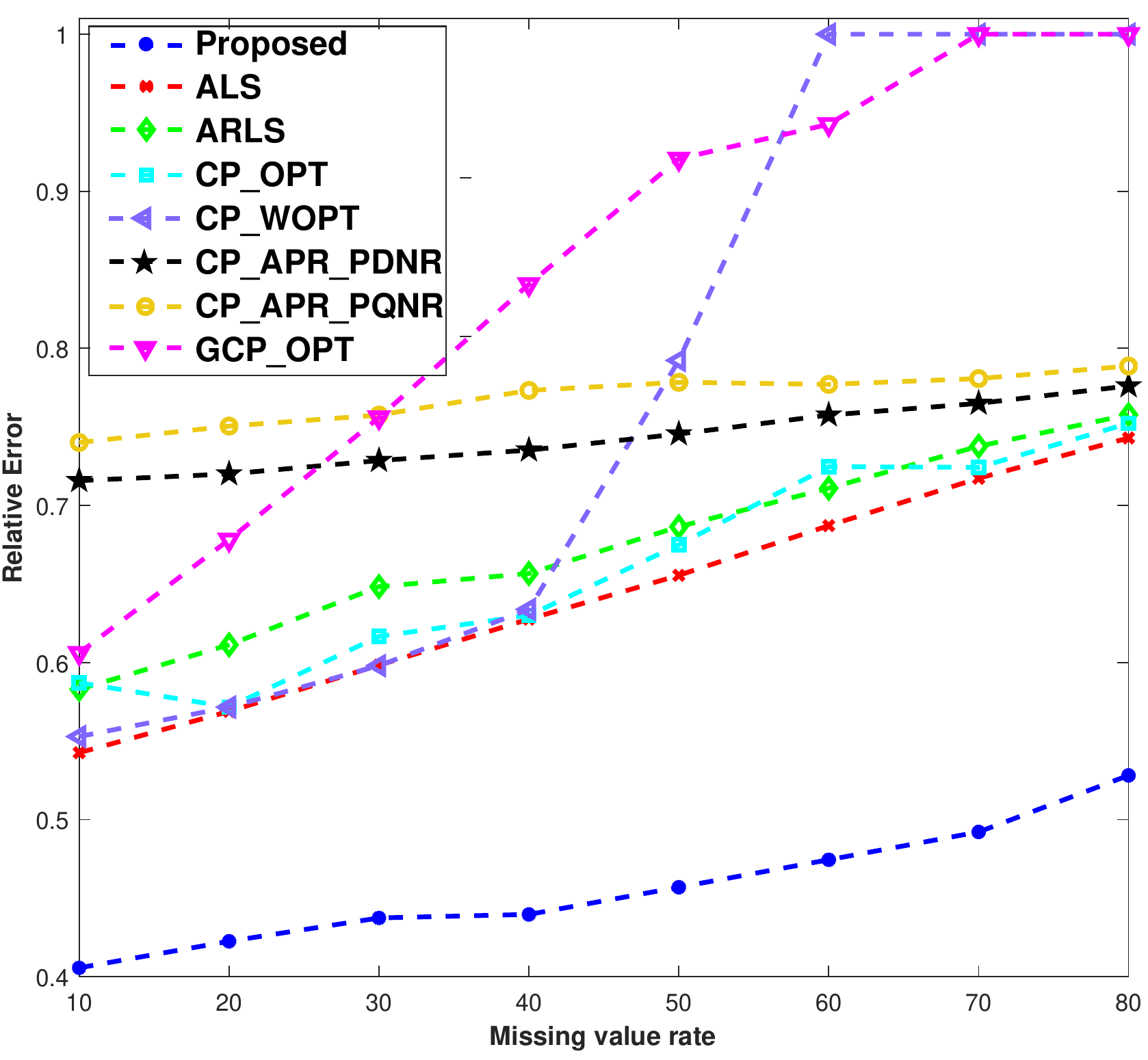}
         \caption{R=5}
         \label{porto_random_r_5}
     \end{subfigure}
        \caption{Recovering Porto traffic tensor with random missing values: Relative Error results}
        \label{porto_random}
\end{figure*}

\begin{figure*}[h!]
     \centering
     \begin{subfigure}[b]{.3\textwidth}
         \centering
         \includegraphics[scale=0.32]{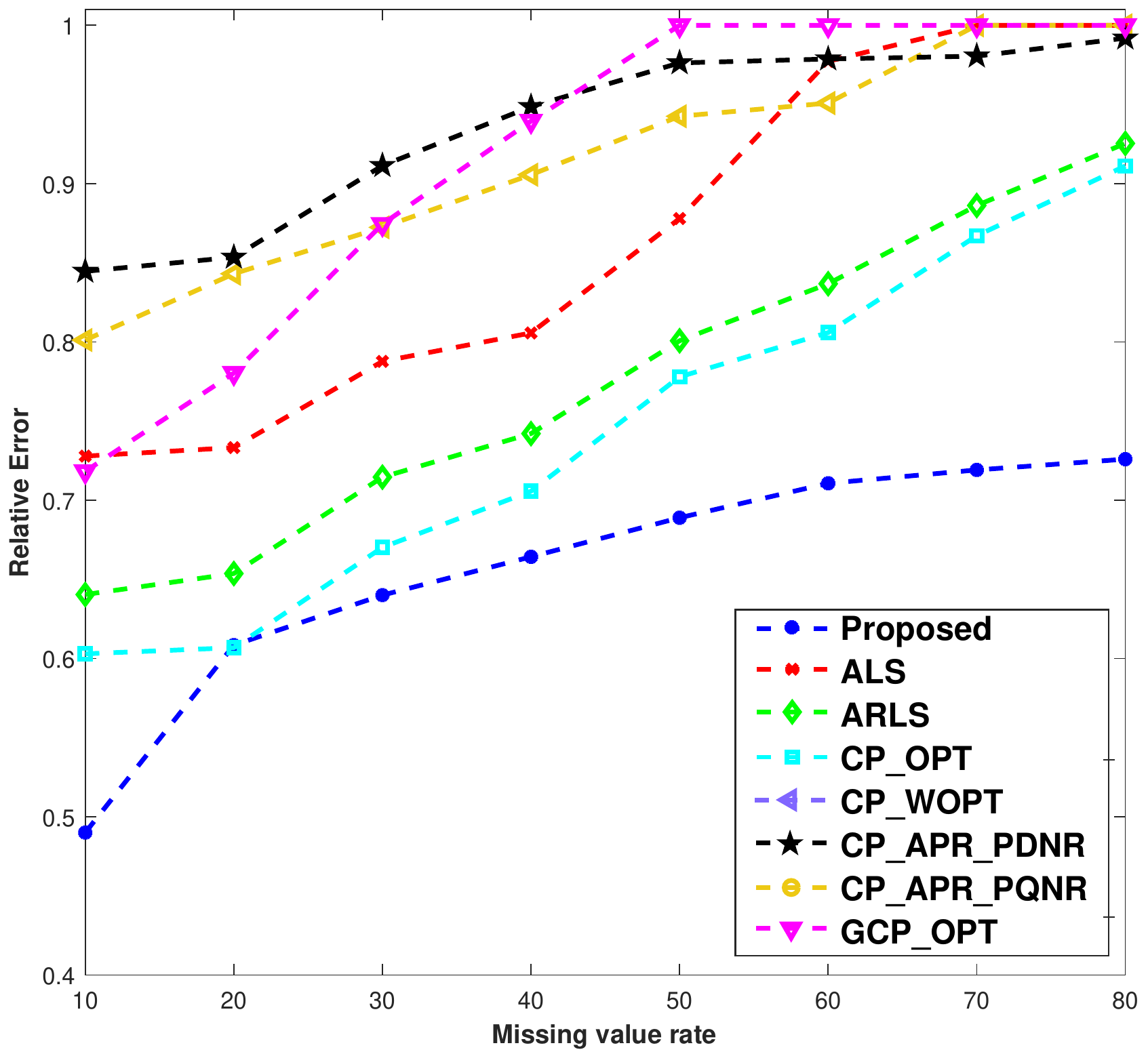}
         \caption{R=3}
         \label{beijing_random_r_3}
     \end{subfigure}
     \hfill
     \begin{subfigure}[b]{0.33\textwidth}
         \centering
         \includegraphics[scale=0.315]{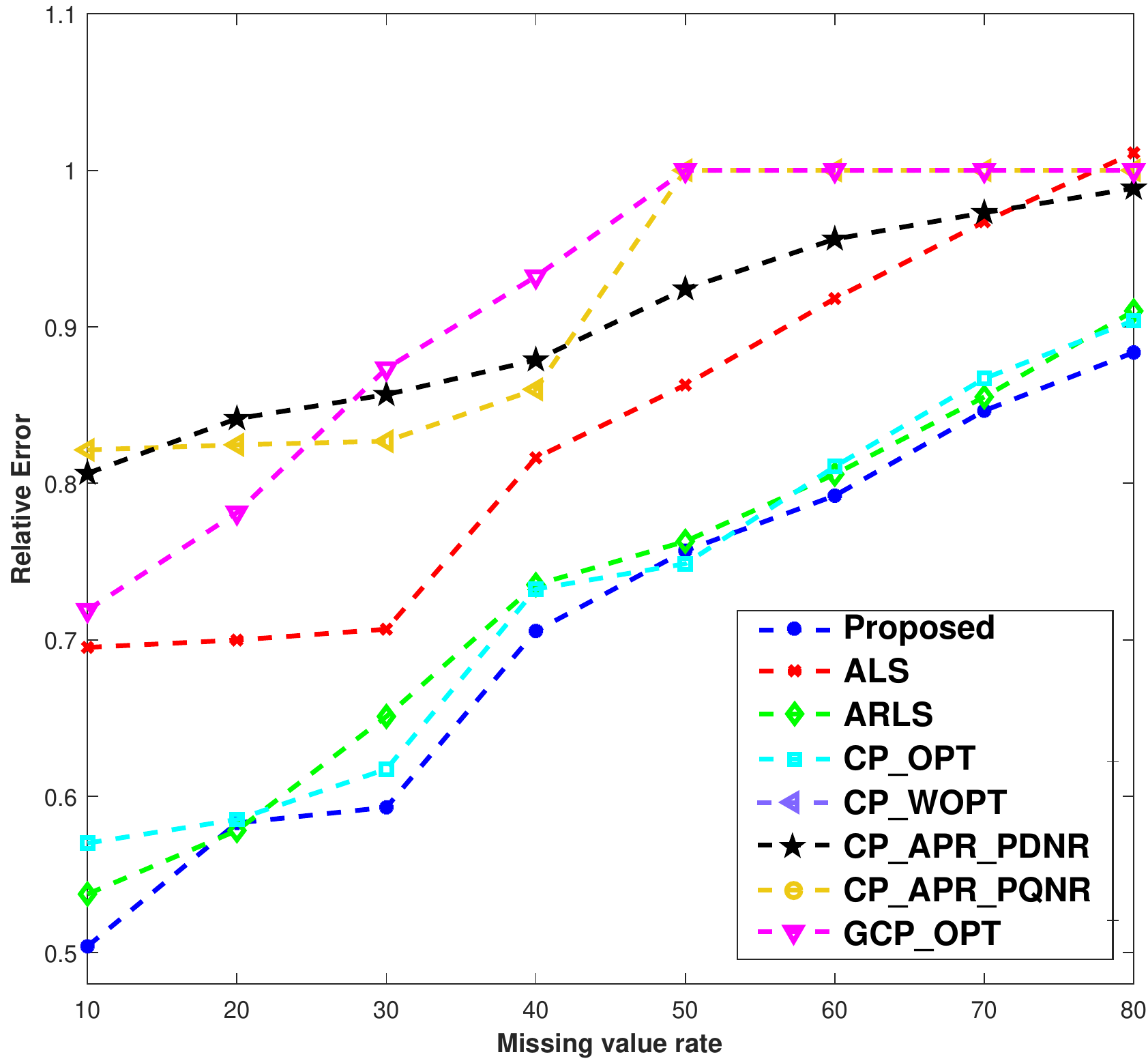}
         \caption{R=4}
         \label{beijing_random_r_4}
     \end{subfigure}
     \hfill
     \begin{subfigure}[b]{0.33\textwidth}
         \centering
         \includegraphics[scale=0.32]{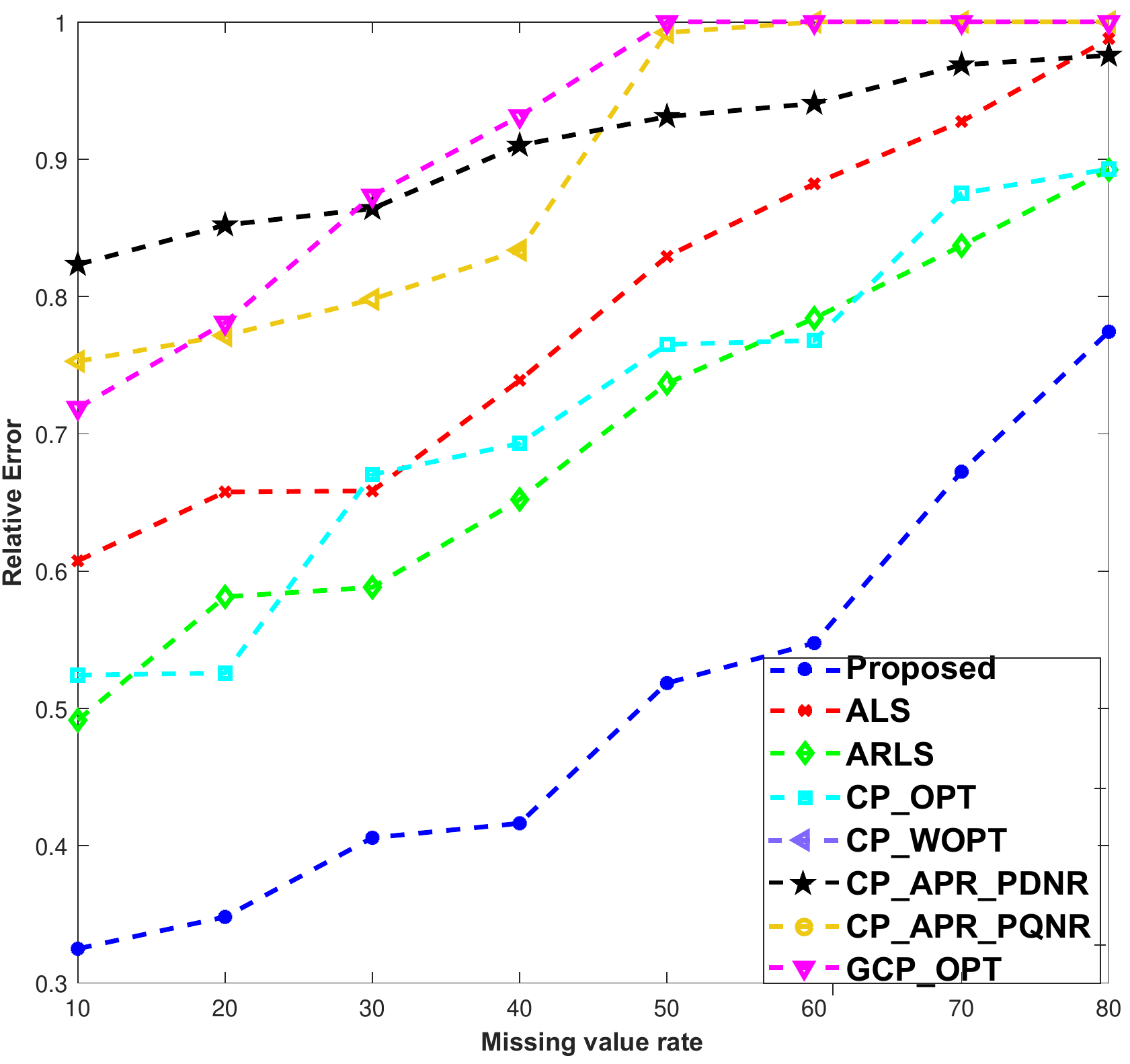}
         \caption{R=5}
         \label{beijing_random_r_5}
     \end{subfigure}
        \caption{Recovering Beijing traffic tensor with random missing values: Relative Error results}
        \label{beijing_random}
\end{figure*}

\begin{figure*}[h!]
     \centering
     \begin{subfigure}[b]{.49\textwidth}
         \centering
         \includegraphics[scale=0.66]{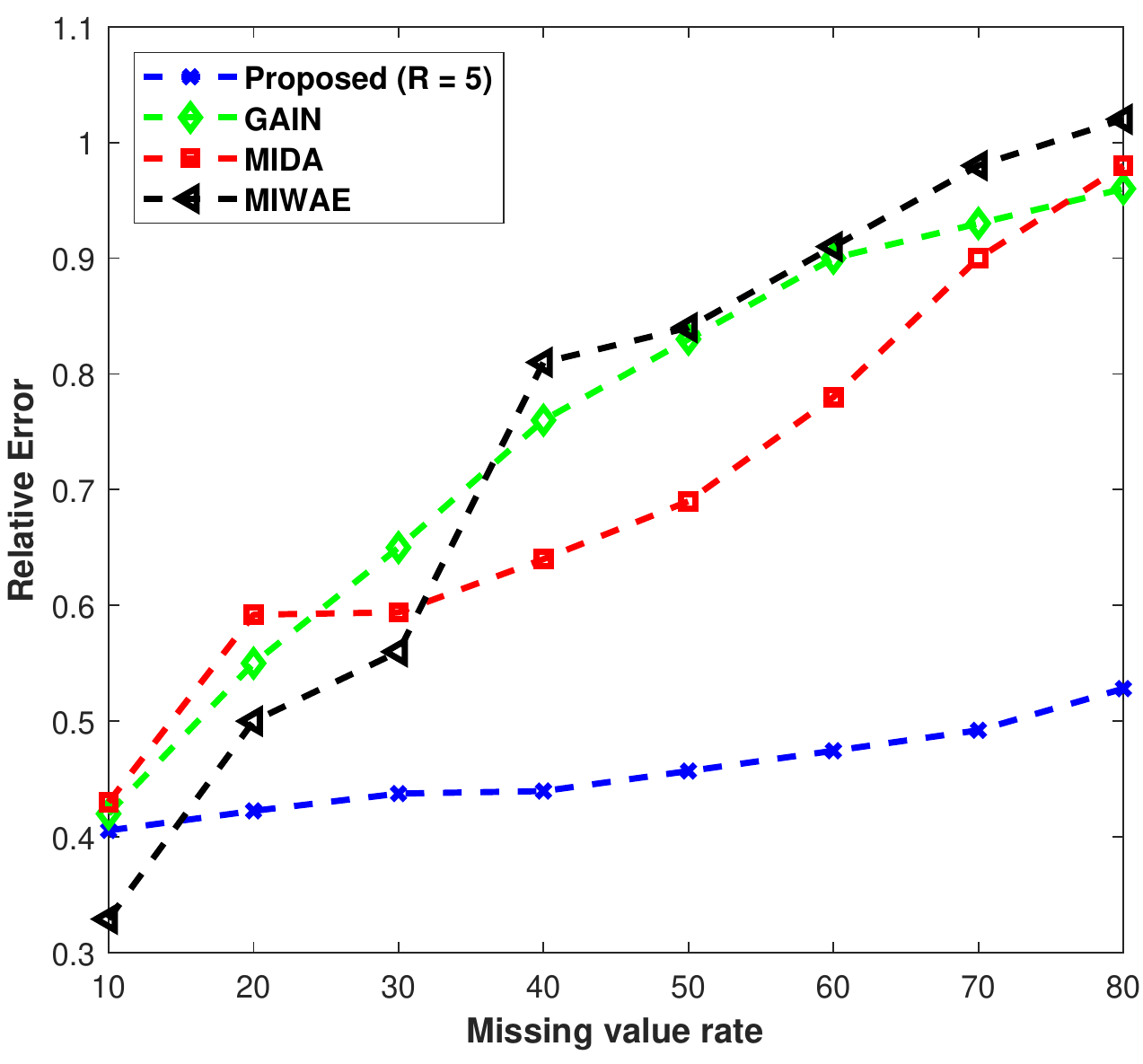}
         \caption{Porto traffic data}
         \label{beijing_random_r_3}
     \end{subfigure}
     \hfill
     \begin{subfigure}[b]{0.49\textwidth}
         \centering
         \includegraphics[scale=0.68]{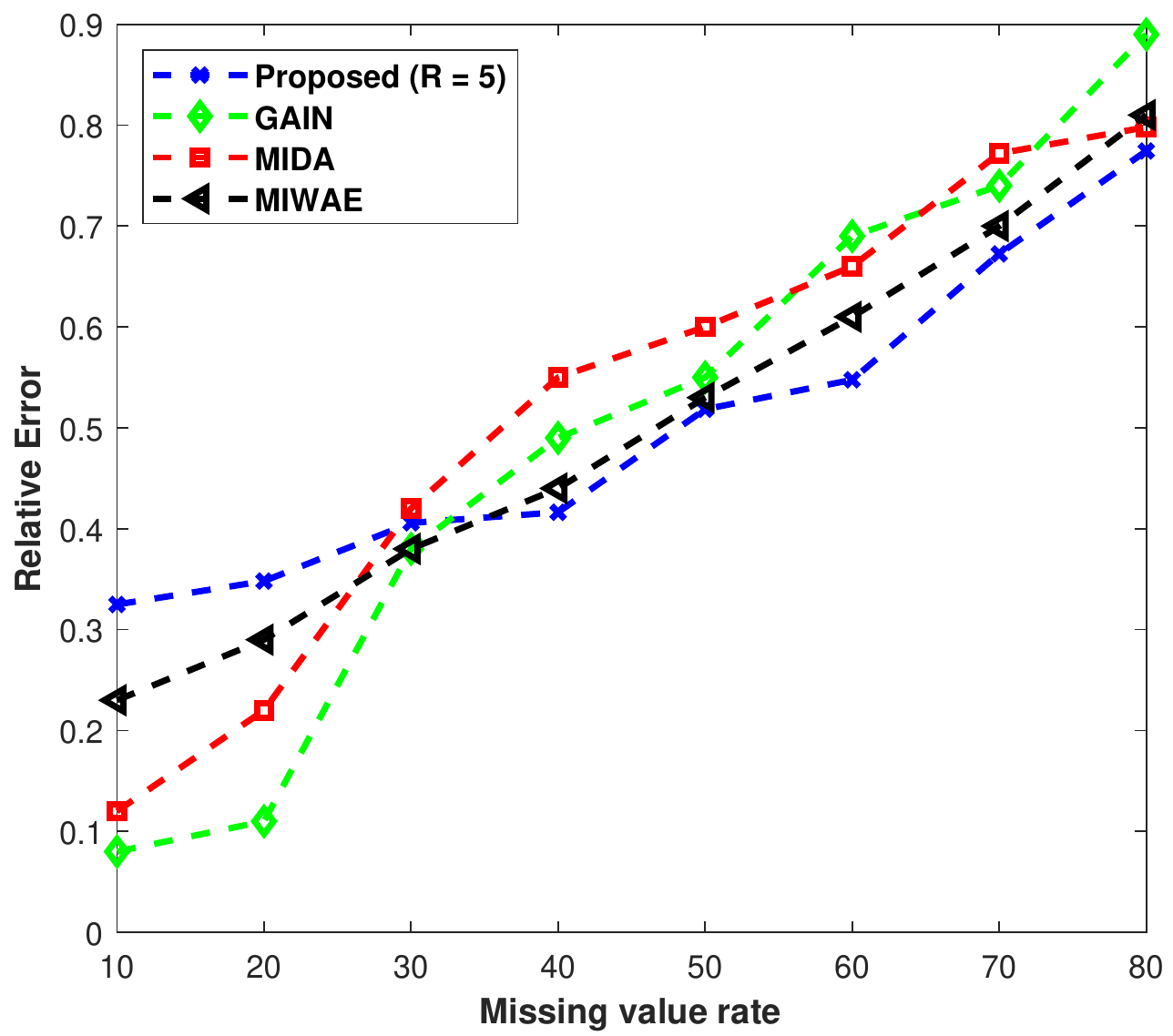}
         \caption{Beijing traffic data}
         \label{beijing_random_r_4}
     \end{subfigure}
        \caption{Comparison with generative models}
        \label{gan_random}
\end{figure*}

We illustrate in Fig. \ref{porto_random} the variation of RE with respect to varying missing values rate for Porto Taxi. We run this simulation using the regularization parameters $\lambda=\beta=0.1$. Results show that the proposed tensor completion approach achieved the best performance for low and high missing value rates and varying $R$. We notice 23\% improvement compared to the closest performance for $R=4$ with low rate missing values. We notice that GCP\_OPT and CP\_WOPT completely fail to recover data for high missing value rates. Figure \ref{beijing_random} illustrates the recovery performance of Beijing T-drive Taxi data with random missing values. The findings confirm the effectiveness of our approach in completing the traffic tensor with 33\% improvement compared to the closest performance for $R=5$ and low missing values rate. We notice  that for high missing rates, GCP\_OPT, CP\_WOPT could not recover the tensor data.
\newline
We illustrate in Fig. \ref{gan_random} the comparison of the proposed completion approach against state-of-art generative models: GAIN, MIDA and MIWAE. We report in this comparison the best performance achieved by the proposed technique. The results show that for low missing rates these models achieved better performance. However, for severe missing rates, the proposed approach achieved significantly better performance with 30\% improvement for Porto traffic tensor having 80\% missing rate.
\subsection{Recovering tensor with structured missing values}
\begin{figure*}[h!]
     \centering
     \begin{subfigure}[b]{.3\textwidth}
         \centering
         \includegraphics[scale=0.32]{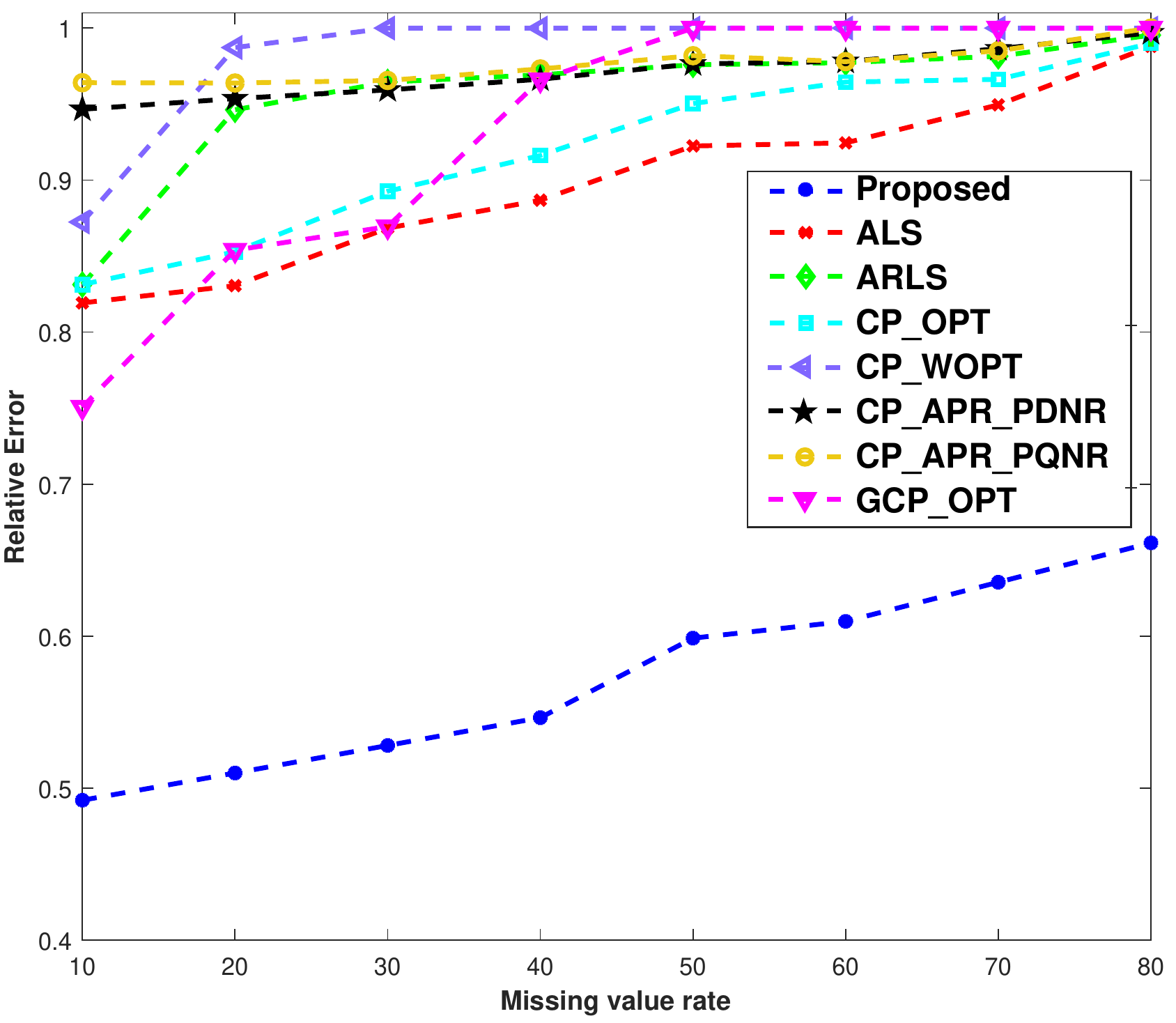}
         \caption{R=3}
         \label{porto_st_r_3}
     \end{subfigure}
     \hfill
     \begin{subfigure}[b]{0.33\textwidth}
         \centering
         \includegraphics[scale=0.315]{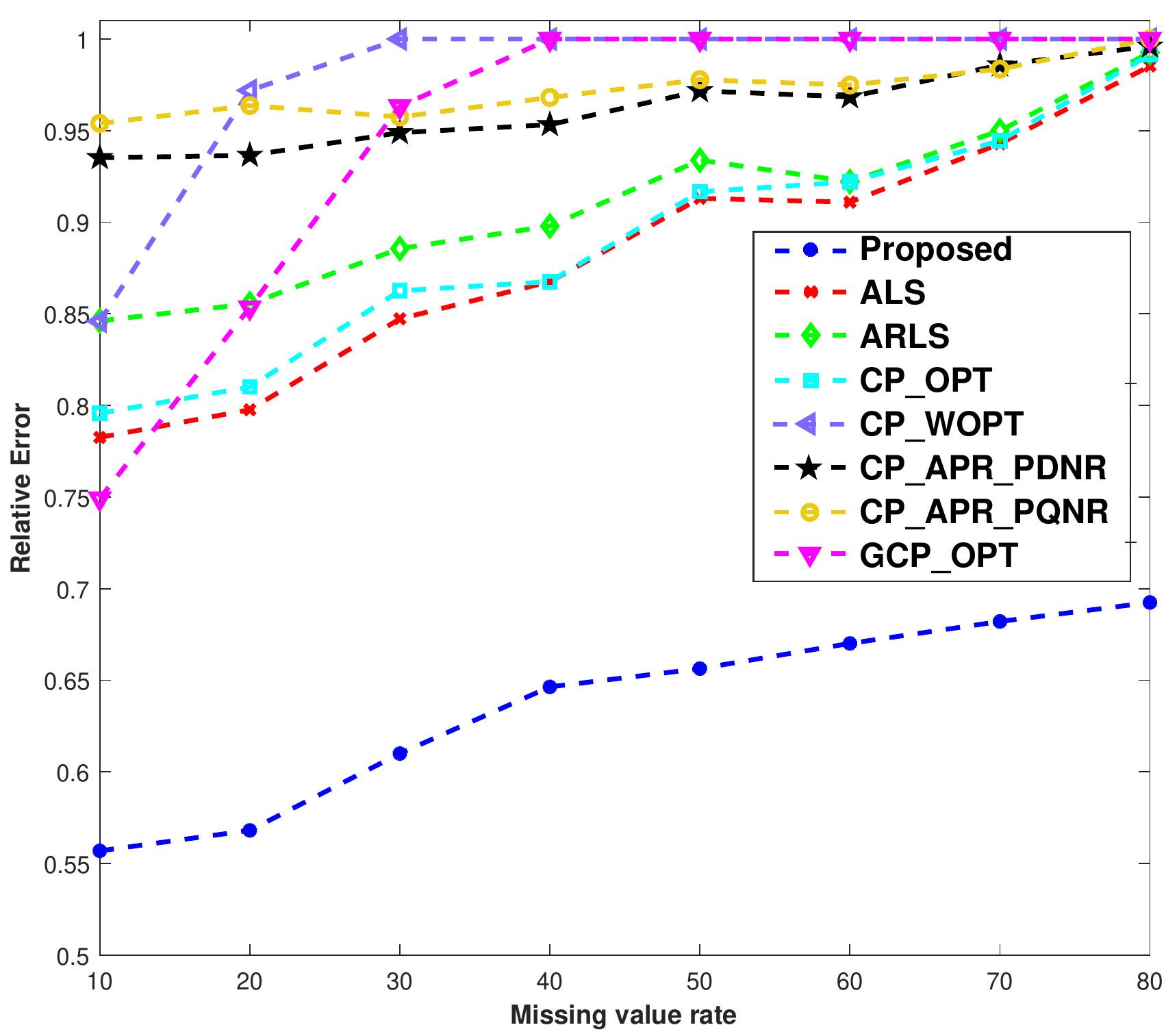}
         \caption{R=4}
         \label{porto_st_r_4}
     \end{subfigure}
     \hfill
     \begin{subfigure}[b]{0.33\textwidth}
         \centering
         \includegraphics[scale=0.315]{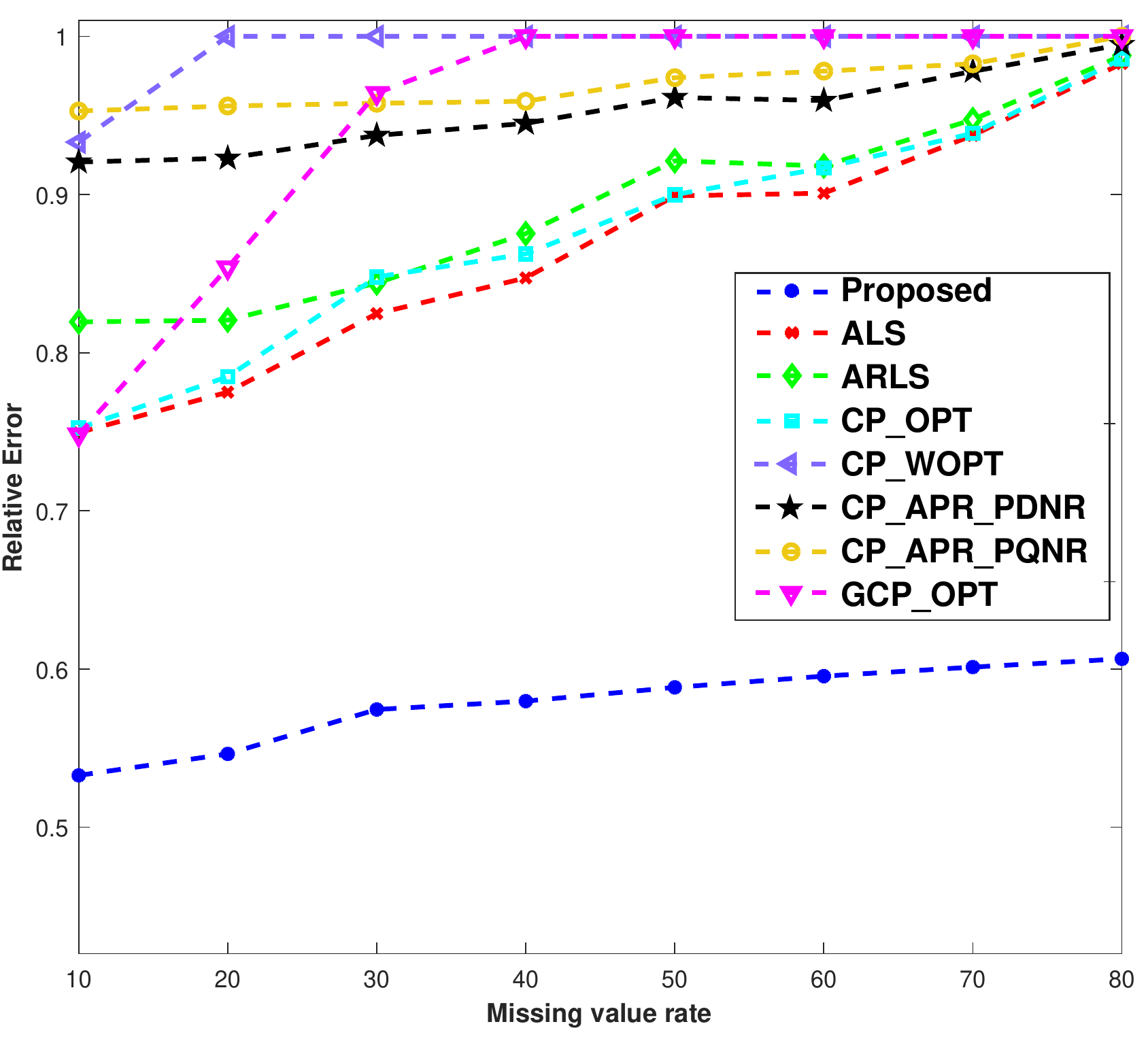}
         \caption{R=5}
         \label{porto_st_r_5}
     \end{subfigure}
        \caption{Recovering Porto traffic tensor with structured loss of values: Relative Error results}
        \label{porto_st}
\end{figure*}

\begin{figure*}[h!]
     \centering
     \begin{subfigure}[b]{.3\textwidth}
         \centering
         \includegraphics[scale=0.32]{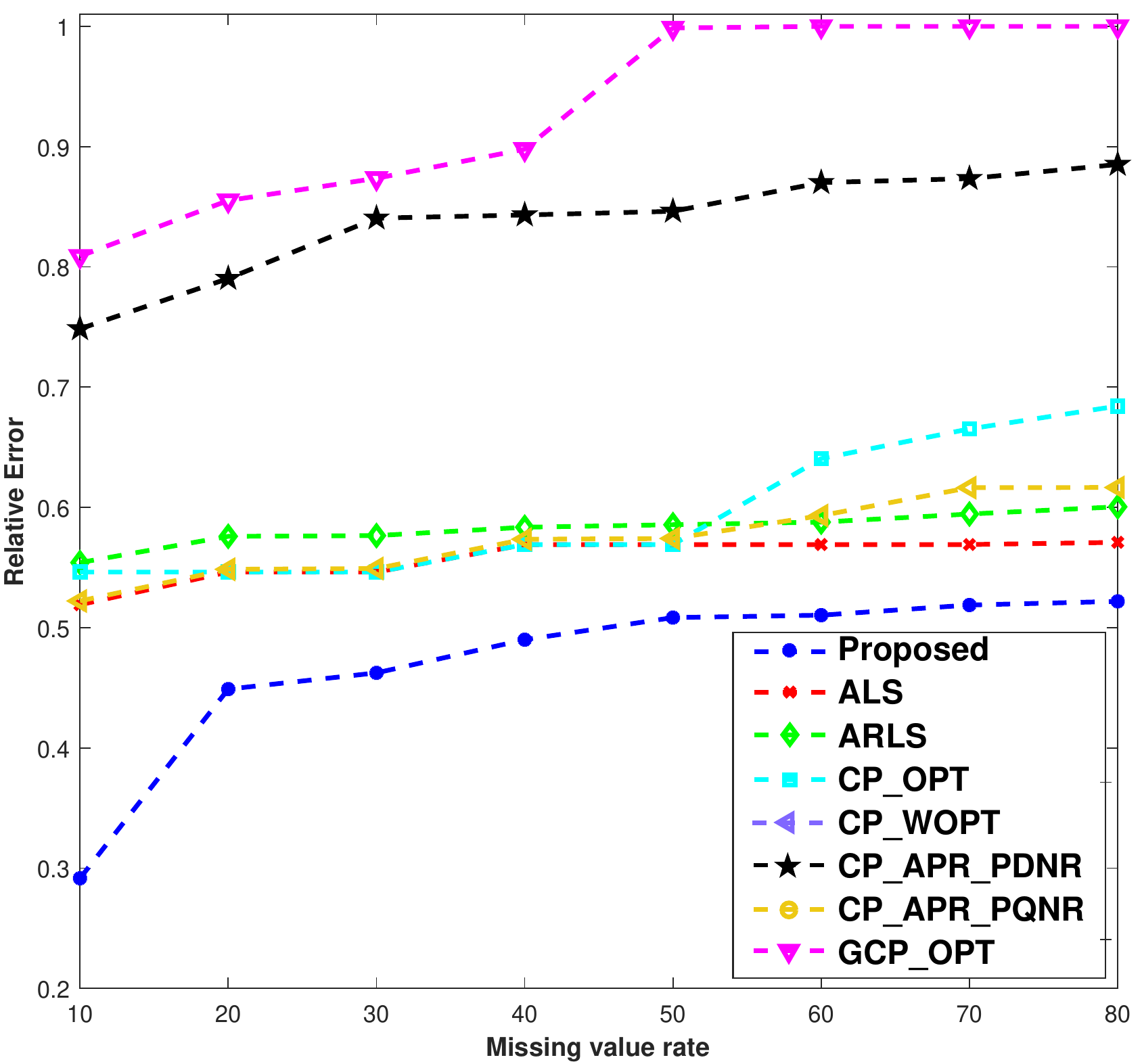}
         \caption{R=3}
         \label{beijing_st_r_3}
     \end{subfigure}
     \hfill
     \begin{subfigure}[b]{0.33\textwidth}
         \centering
         \includegraphics[scale=0.315]{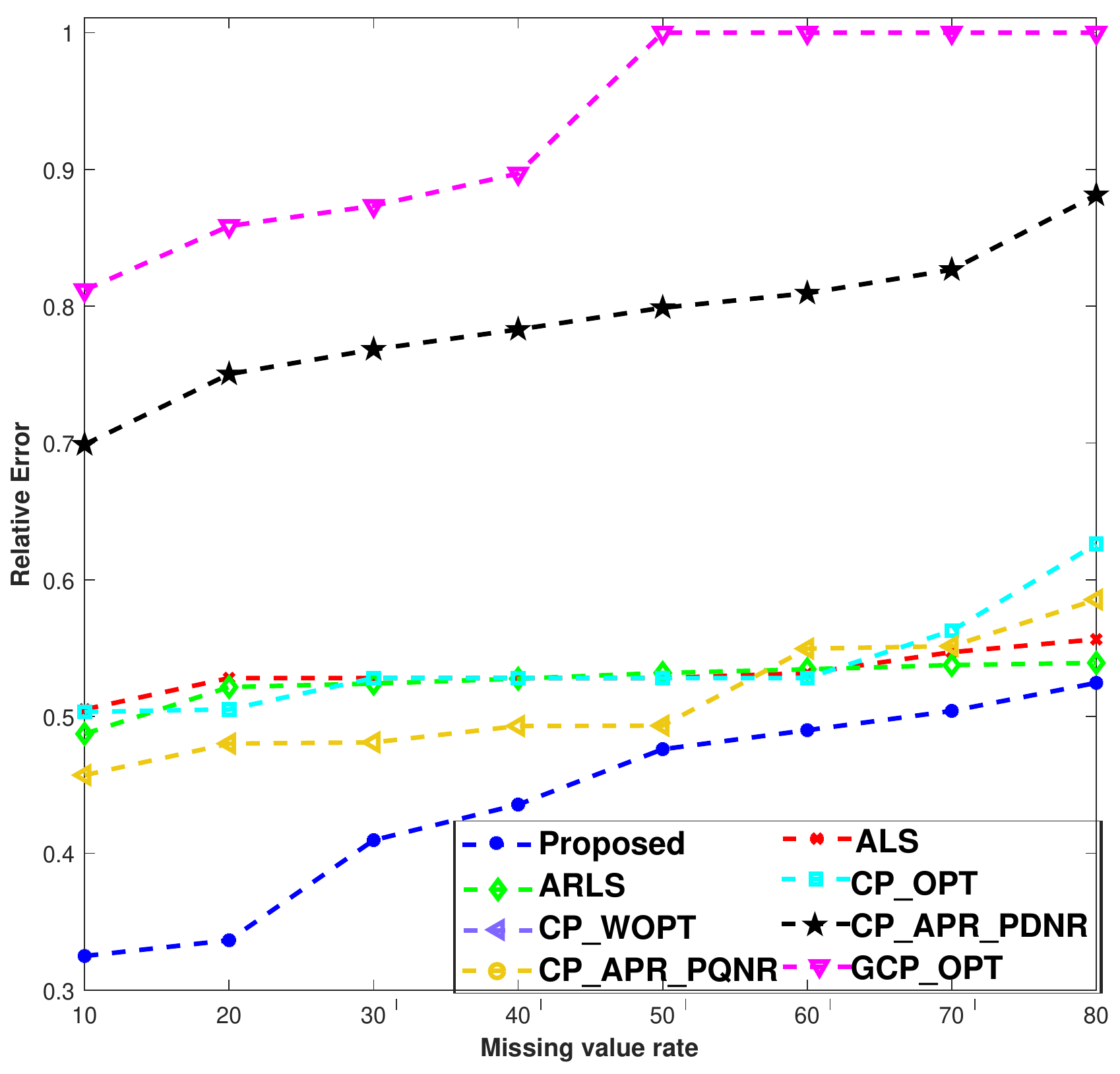}
         \caption{R=4}
         \label{beijing_st_r_4}
     \end{subfigure}
     \hfill
     \begin{subfigure}[b]{0.33\textwidth}
         \centering
         \includegraphics[scale=0.315]{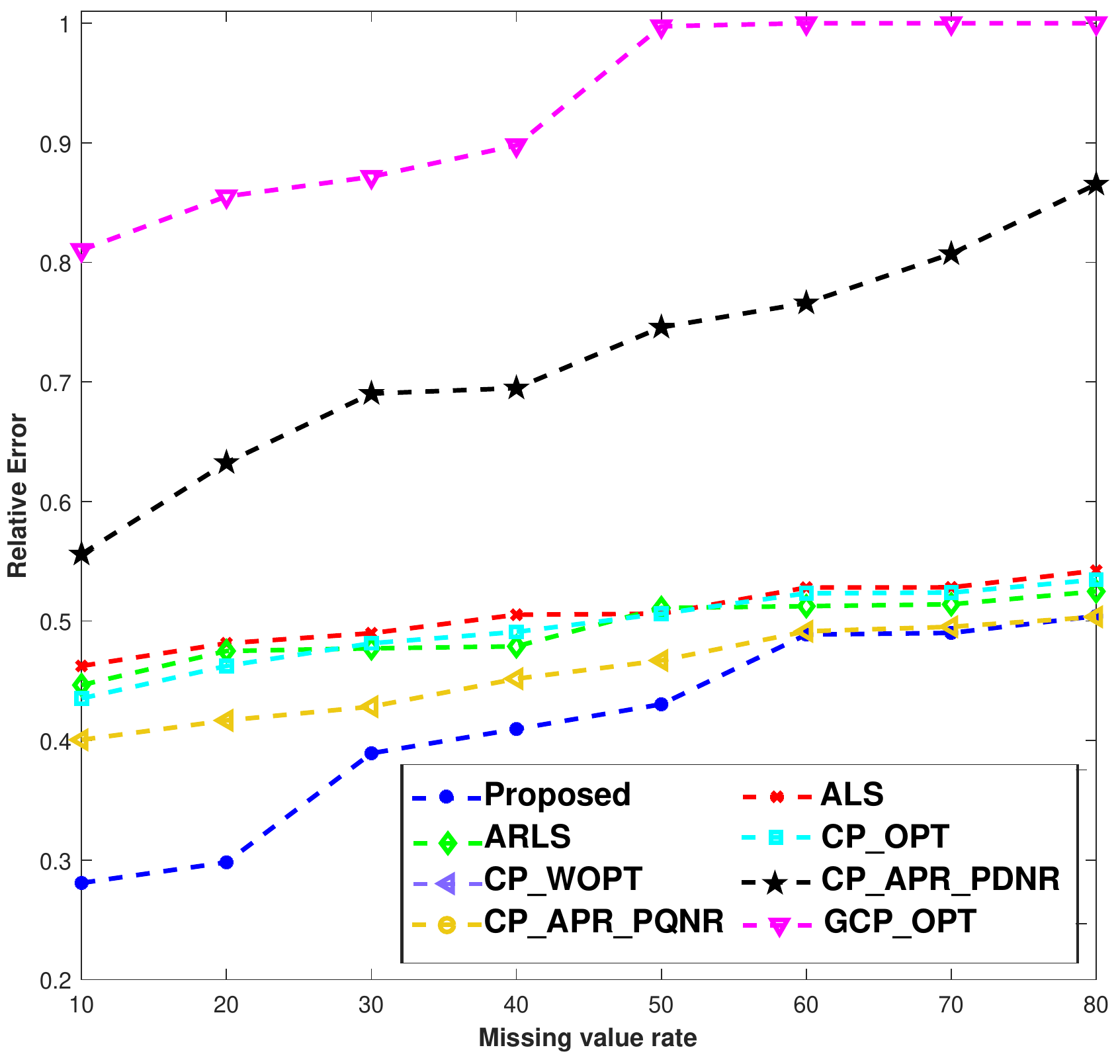}
         \caption{R=5}
         \label{beijing_st_r_5}
     \end{subfigure}
        \caption{Relative Error results for recovering Beijing traffic tensor with structured missing values}
        \label{beijing_st}
\end{figure*}

\begin{figure*}[h!]
     \centering
     \begin{subfigure}[b]{.49\textwidth}
         \centering
         \includegraphics[scale=0.62]{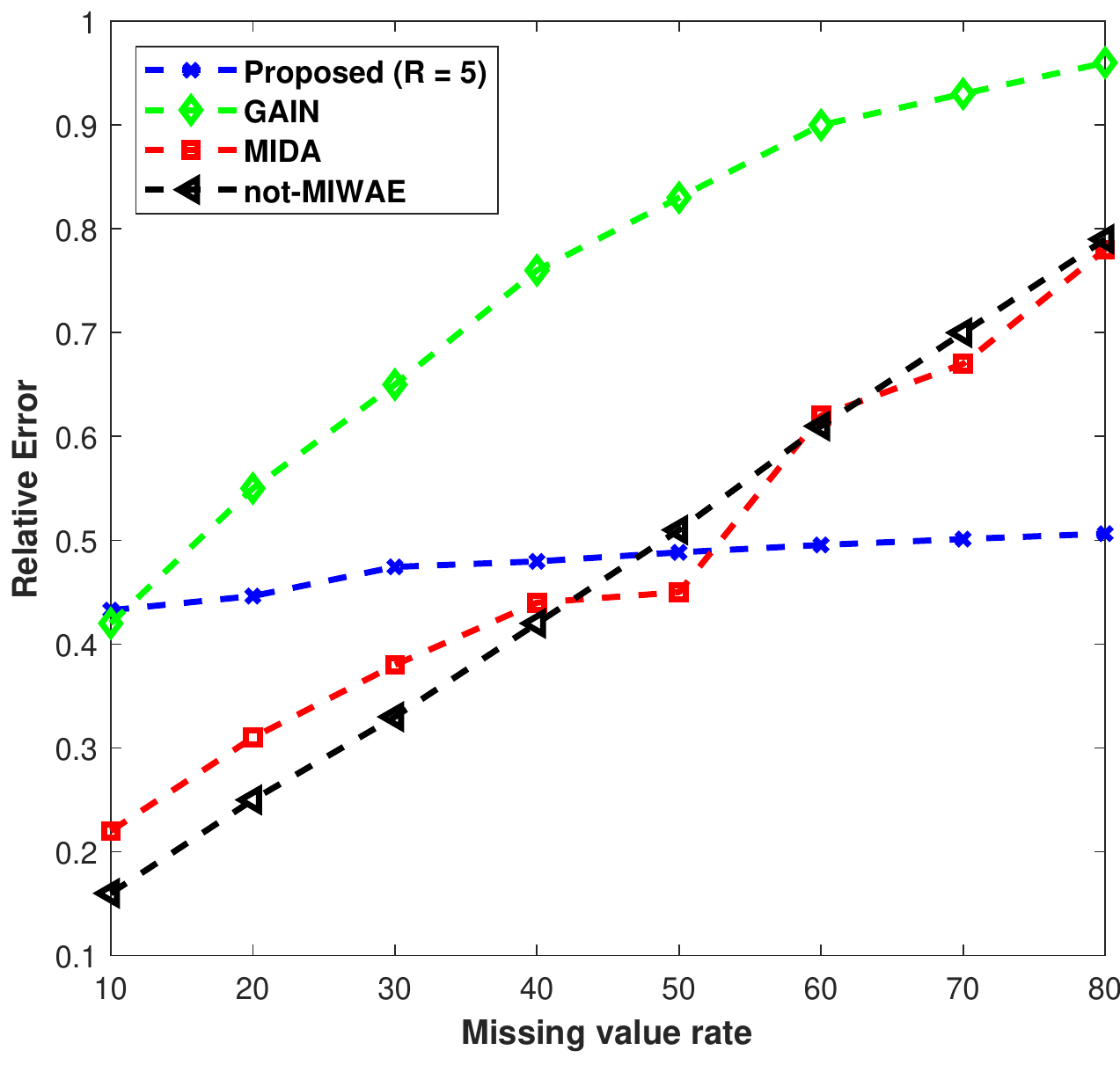}
         \caption{Porto traffic data}
         \label{beijing_random_r_3}
     \end{subfigure}
     \hfill
     \begin{subfigure}[b]{0.49\textwidth}
         \centering
         \includegraphics[scale=0.68]{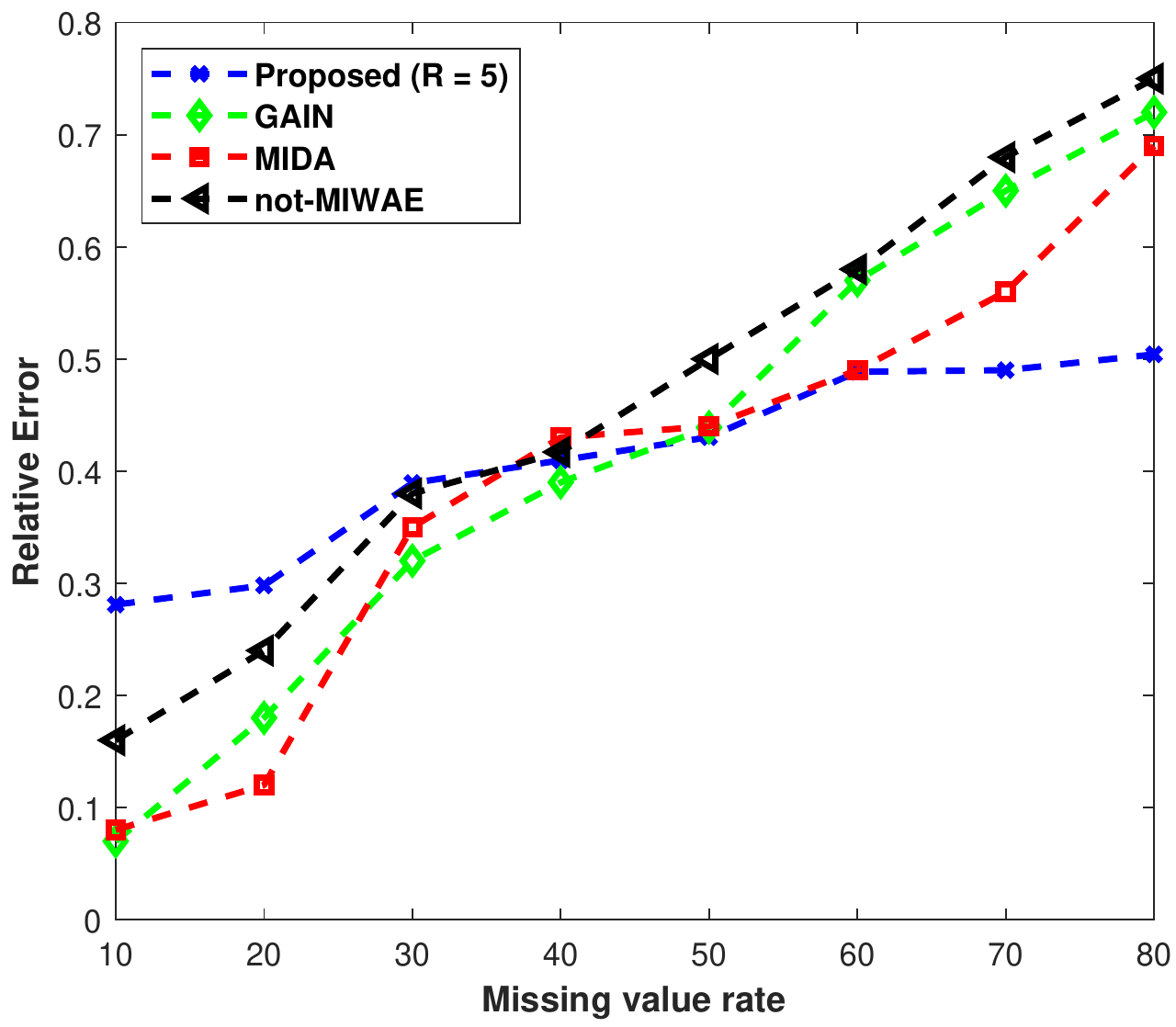}
         \caption{Beijing traffic data}
         \label{beijing_random_r_4}
     \end{subfigure}
        \caption{Comparison with generative models and structured missing values}
        \label{gan_st_loss}
\end{figure*}

In this simulation scenario, we set $\lambda=0.1$ and $\beta=0.01$.
Figure \ref{porto_st} depicts the RE results for different settings. The best recovery performance is achieved by our method while GCP\_OPT results in high RE. Most of the algorithms showed stable performance except CP\_APR\_PQNR and GCP\_OPT. CP\_WOPT again results in similar performance as in the previous scenario. 
We illustrate in Fig. \ref{beijing_st} the recovery performance for Beijing T-drive taxi with structured missing values. Our  CP completion approach achieved the lowest RE values for all missing values rate with 26\% improvement compared to the closest performance for $R=3$ and low missing values rate. Fig. \ref{gan_st_loss} depicts the comparison of the proposed completion method against GAIN, MIDA and not-MIWAE. Note that not-MIWAE is designed to recover data with structured missing values. The findings show similar performance to the random missingness experiment. In fact, although the generative models achieved better performance for low missing rates, a performance degradation is witnessed for higher missing rates. For 80\% missing rate, our method exhibited an improvement of 35\% compared to the closest performance for Porto traffic data.
\subsection{Discussion}
\begin{figure}[h!]
  \centering
  \includegraphics[scale=0.4]{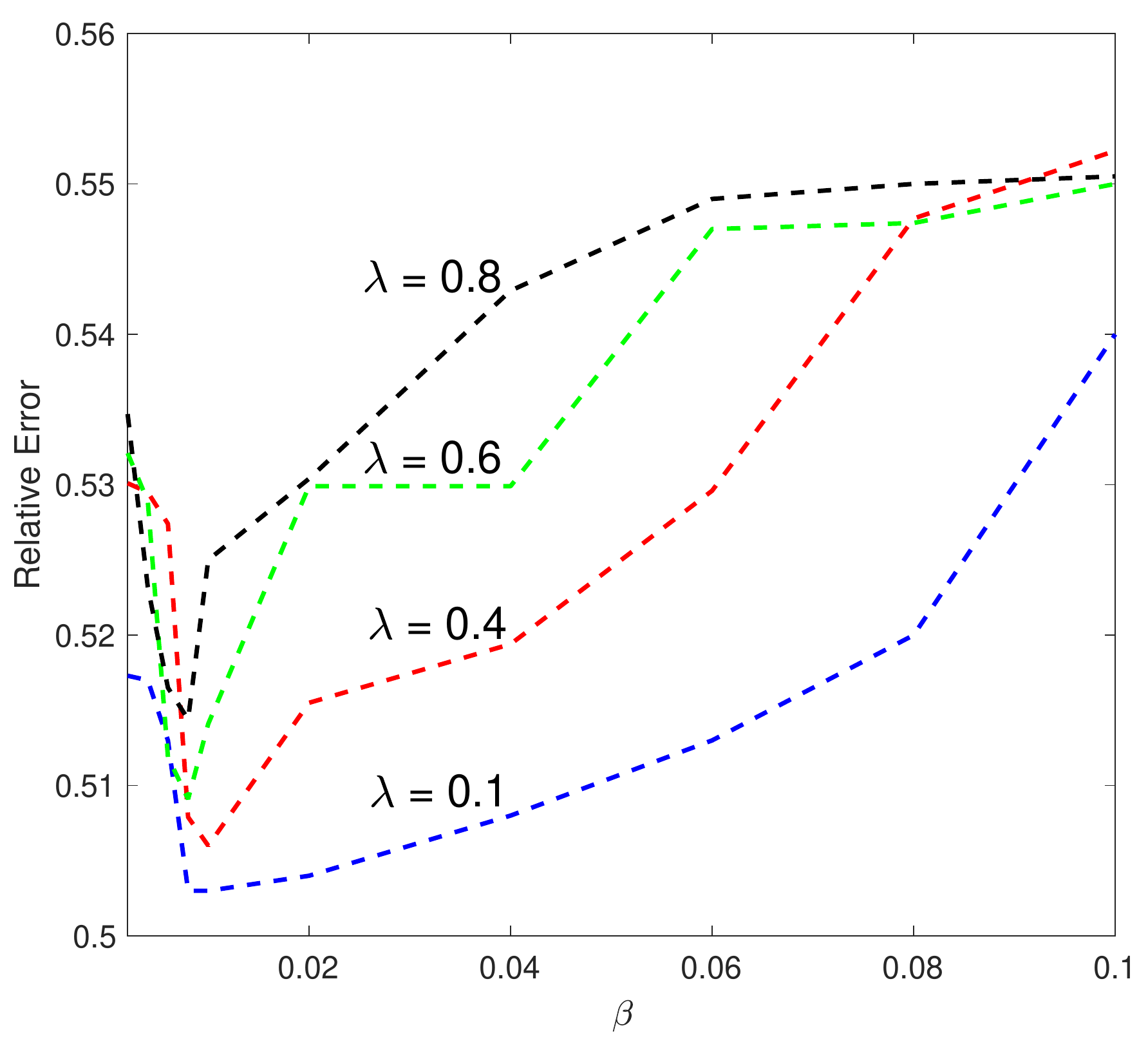}
  \caption{Recovering Porto traffic data under structured missing values: Variation of the Relative Error with respect to $\lambda$ and $\beta$}
  \label{beta}
\end{figure}

\begin{figure}[h!]
  \centering
  \includegraphics[scale=0.3]{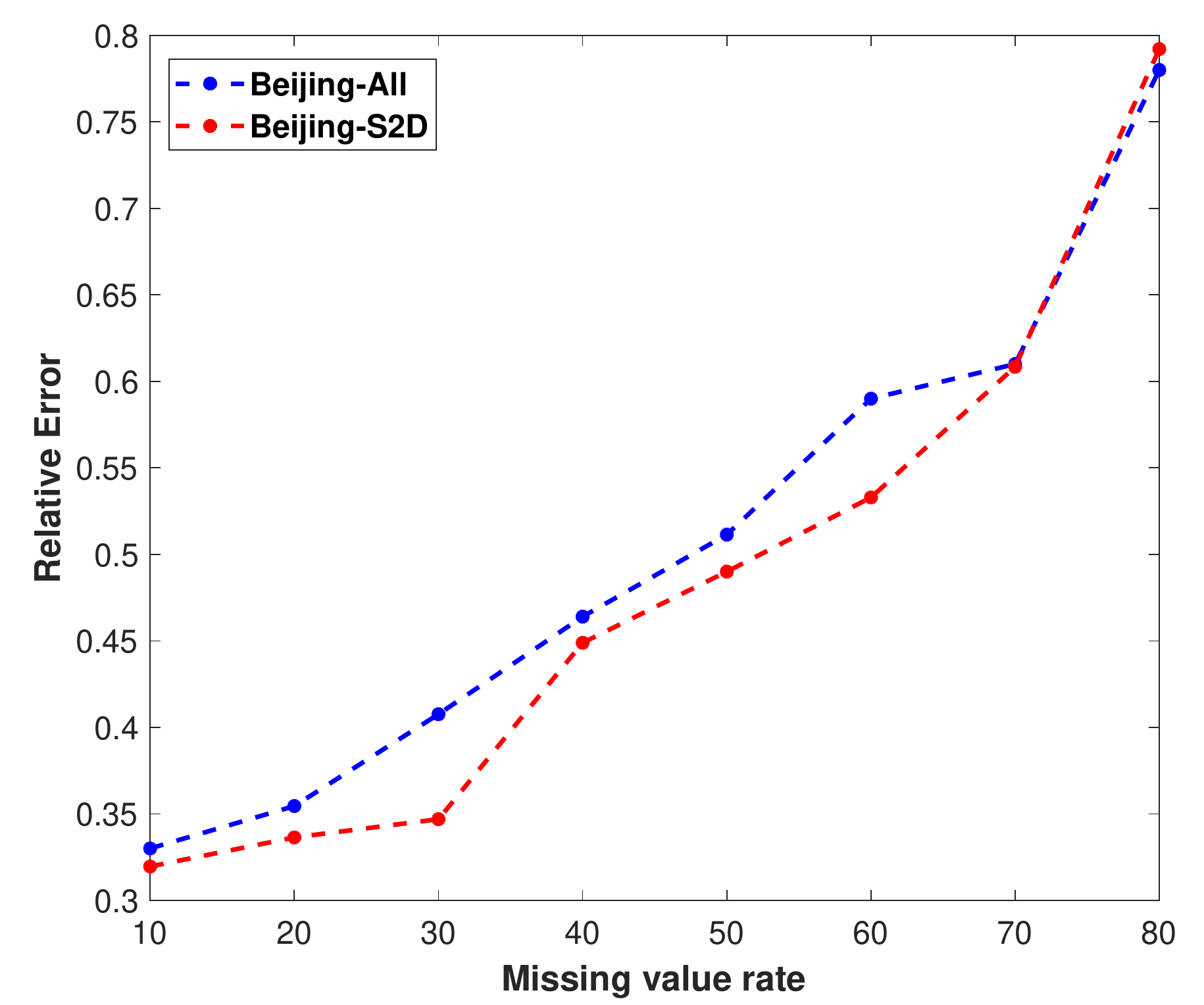}
  \caption{Recovering T-Drive data constructed with source and destination only and all trajectories' locations: Random missing values}
  \label{time}
\end{figure}

\begin{figure}[h!]
  \centering
  \includegraphics[scale=0.3]{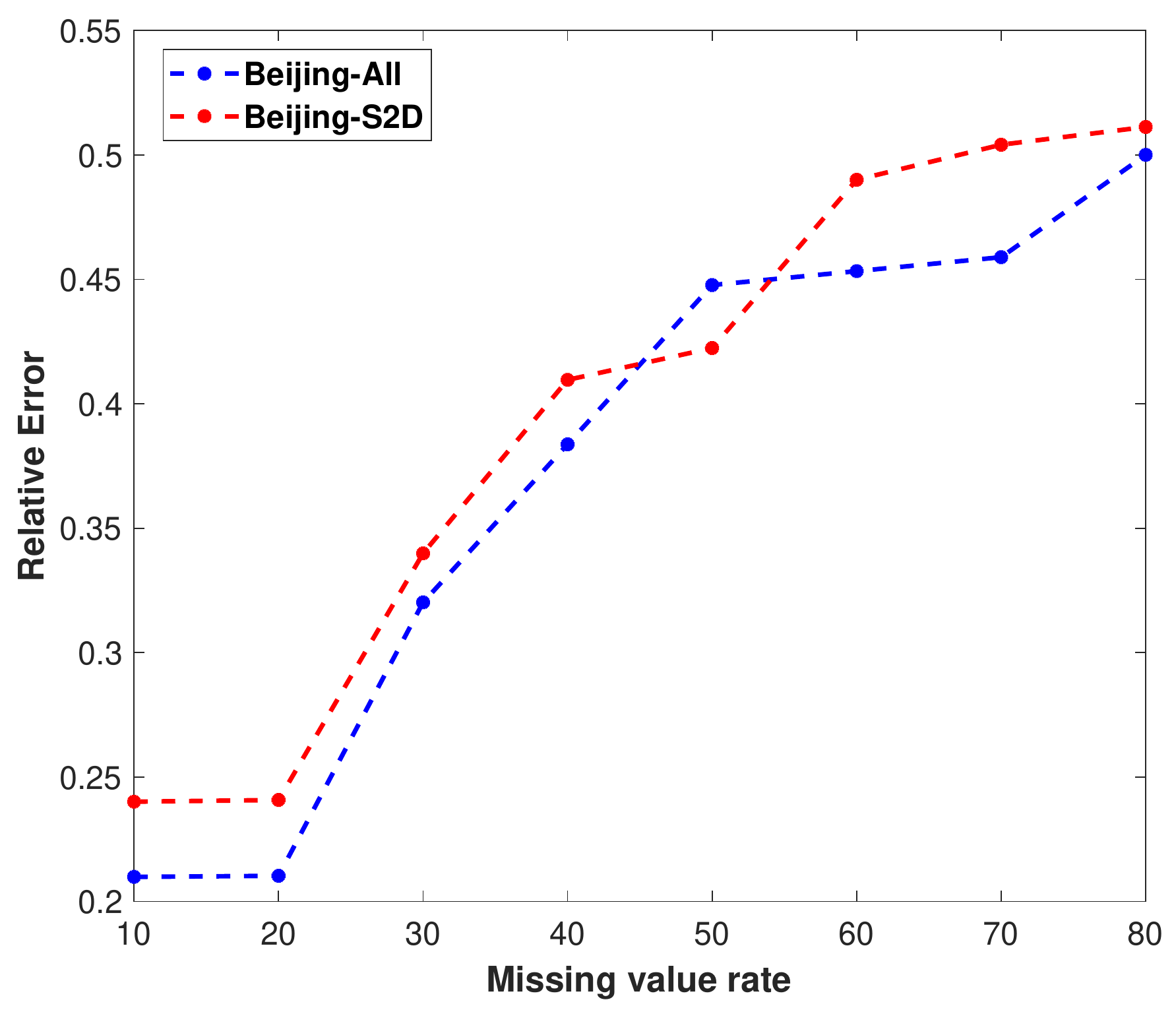}
  \caption{Recovering T-Drive data constructed with source and destination only and all trajectories' locations: Structured missing values}
  \label{dense_sparse1}
\end{figure}

\begin{figure}[h!]
  \centering
  \includegraphics[scale=0.5]{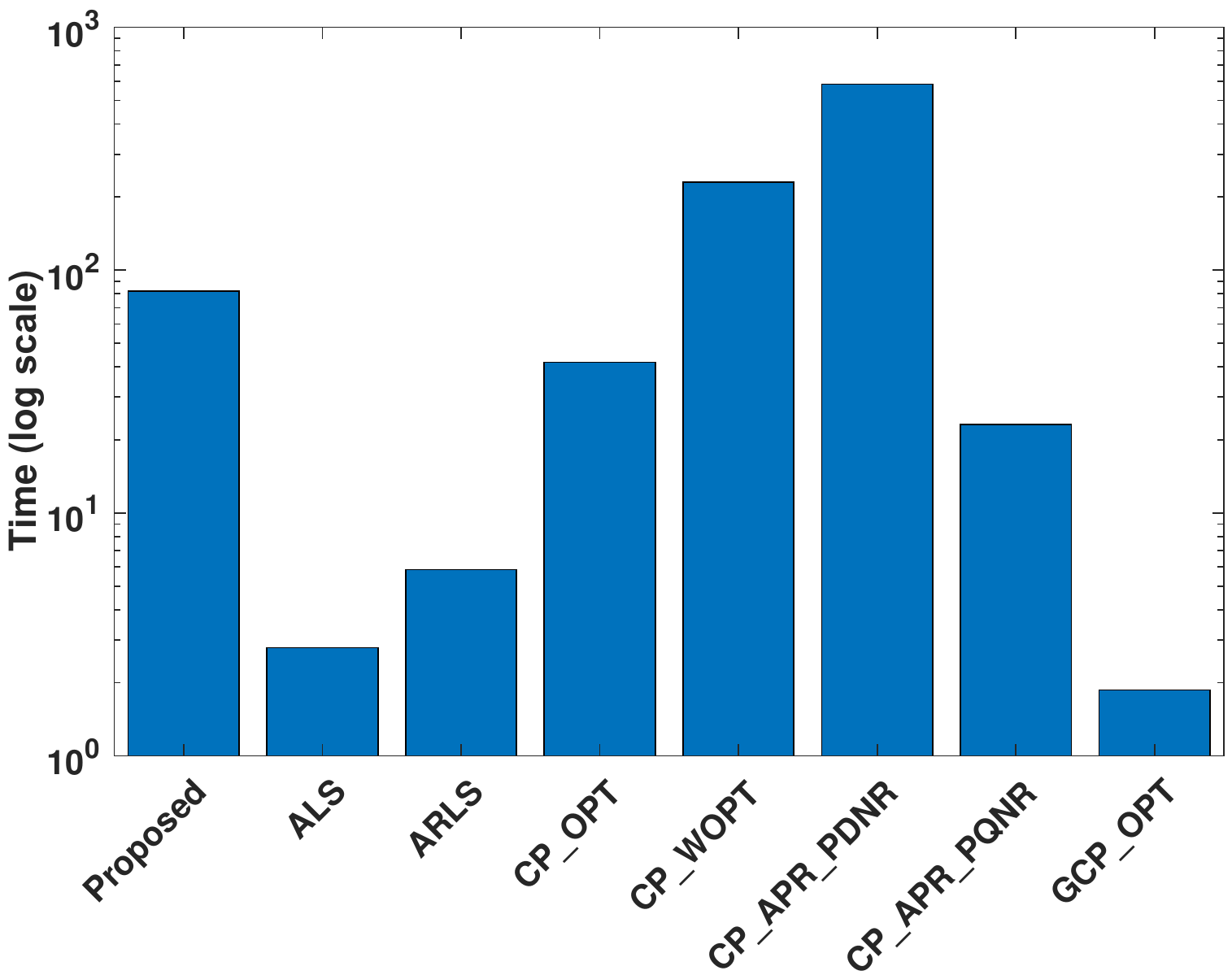}
  \caption{Time complexity}
  \label{dense_sparse2}
\end{figure}

Experiments showed that our urban and time aware CP tensor completion approach is efficient in recovering missing traffic information with both random and structured missing values. The other algorithms, although achieved competitive performance with low missing values rate, they failed to recover the traffic tensor with very high imputation. On overall, the proposed technique performed better on T-Drive data compared to Porto Taxi. This has been also the case for all techniques used for comparison. 
\newline 
\indent It is worth noting that performance of our approach depends on the regularization parameter $\lambda$ and $\beta$. We analyze the variation of the Relative Error with respect to different $\lambda$ and $\beta$ to recover Porto data under structured missing values. The results are depicted in Fig. \ref{beta}. We notice that high regularization of the matrix norm, i.e. high $\lambda$, value leads to higher error. In addition, we notice that the optimal choice for parameter $\beta$ is in the range of 0.01 to 0.02. Higher or lower values result in higher RE. Automatic tuning of these parameters is an open research question. For this work, $\lambda$ and $\beta$ parameters are empirically chosen. 
\newline 
\indent  We further compare the performance on tensors constructed using the source and destination sub-regions only and using all locations visited in the journey from source to destination. Without loss of generality, we conduct this experiment  on T-Drive data with $R=6$ where we attempt to recover the tensors under different corruption levels with random and structure missing values and evaluate the $RE$. We refer to the first tensor as Beijing-S2D and the second one as Beijing-All.
To quantify the sparsity of each tensor, we use the $S_{\frac{l_2}{l_1}}$ \cite{sparsity_measure}:
\begin{equation}
    S_{log}(x) = - \sum_i log(1+x_i^2)
\end{equation}
Computation of the sparsity measure shows that Beijing-S2D is more sparse than Beijing-All with 5 order of magnitude where $S_{log} = -1.2 10^6$  and $S_{log}= -6.5 10^6$  for Beijing-S2D and Beijing-All respectively. Results, depicted in Figures \ref{dense_sparse1} and \ref{dense_sparse2} show that the proposed approach achieved similar performances on both tensors and under both missing values scenarios. Hence, we can conclude that it is not affected by the sparsity of the input tensor. Our design choice for the traffic tensor is motivated by the fact that using all trajectories' locations would result in constructing a traffic tensor that better reflects the traffic information in the area of study.
\newline
\indent Finally, we analyze the time complexity of each completion approach. We run each algorithm until the minimization of its objective function is less than $10^{-6}$. We illustrate in Fig. \ref{time} the results of the experiment with time in log scale. The proposed approach achieved relatively high time complexity as the algorithm requires computing multiple operations and applying Moore-Penrose inverse. Therefore it is important to achieve a tradeoff between performance and execution time. \newline
\section{Conclusion}
We proposed a CP based completion approach to recover the missing values from traffic tensor. We augmented the CP algorithm with additional information related to the urban context of the area of study. This includes several biodiversity-inspired characteristics related to the richness, diversity, concentration and traffic convenience. In addition, we take into account the temporal information by considering the periodicity of the traffic data. We established two comparison scenarios and analyzed the proposed approach from time complexity perspective. Our findings showed that the CP completion approach augmented with the proposed urban and time information achieved competitive recovery performance.
\newline 
In future work, we will focus on alleviating the time complexity. We will also address the choice  of the regularization parameters and propose a solution for automatic tuning
\section*{Acknlowdgement}
This research was made possible by NPRP 9-224-1-049 grant from the Qatar
National Research Fund (a member of The Qatar Foundation).
The statements made herein are solely the responsibility of the authors.

\begin{IEEEbiography}[{\includegraphics[width=1in,height=1.25in,clip,keepaspectratio]{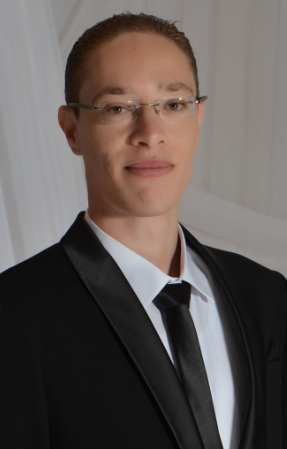}}]{Ahmed Ben Said} received the Ph.D. degree
in computer Science from the University of Burgundy, France, in 2015. He was a Research Assistant with Qatar University on several projects,
including the simulation of a surgical cutting
operation using 3-D modeling, the usage of multispectral image for face recognition, and the development of reliable mHealth system for remote
patient diagnosis. He currently holds a postdoctoral position at Qatar University. His research interests include machine learning and computer vision. He is also interested
in urban computing and mobile health systems.
\end{IEEEbiography}

\begin{IEEEbiography}[{\includegraphics[width=1in,height=1.25in,clip,keepaspectratio]{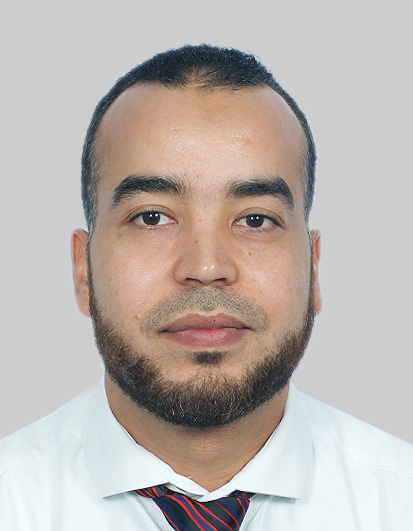}}]
{Abdelkarim Erradi} is an Associate Professor in the Computer Science and Engineering Department at Qatar University. His research and development activities and interests focus on service-oriented computing, cloud Services composition and mobile crowdsensing. He leads several funded research projects in these areas. He has authored several scientific papers in international conferences and journals. He received his Ph.D. in computer science from the University of New South Wales, Sydney, Australia. Besides his academic experience, he possesses 12 years professional experience as a Designer and a Developer of large scale enterprise applications.
\end{IEEEbiography}

\end{document}